\documentclass[runningheads]{llncs}
\usepackage[T1]{fontenc}
\usepackage{graphicx}
\usepackage{xcolor}
\usepackage{soul}
\usepackage{amsmath}
\usepackage{booktabs}
\usepackage{multirow}
\usepackage{float}
\usepackage{orcidlink}

\usepackage{hyperref}
\usepackage{academicons}

\begin{document}
%

\title{Dataset Optimization for Chronic Disease Prediction with Bio-Inspired Feature Selection}
\titlerunning{Optimizing Chronic Disease Prediction}
%
\author{\href{https://arxiv.org/search/cs?searchtype=author&query=Dyoub,+A}{Abeer Dyoub}\inst{1}\orcidlink{0000-0003-0329-2419} \and
\href{https://arxiv.org/search/cs?searchtype=author\&query=Letteri,+I}{Ivan Letteri}\inst{2}\orcidlink{0000-0002-3843-386X} }
\authorrunning{Abeer Dyoub et al.}
%
\institute{University of L'Aquila, Italy \\
\email{abeer.dyoub@univaq.it}\\
\and
Sapienza University of Rome, Italy\\
\email{ivan.letteri@uniroma1.it}}
\maketitle              
\begin{abstract}
In this study, we investigated the application of bio-inspired optimization algorithms, including Genetic Algorithm, Particle Swarm Optimization, and Whale Optimization Algorithm, for feature selection in chronic disease prediction. The primary goal was to enhance the predictive accuracy of models streamline data dimensionality, and make predictions more interpretable and actionable.

The research encompassed a comparative analysis of the three bio-inspired feature selection approaches across diverse chronic diseases, including diabetes, cancer, kidney, and cardiovascular diseases. Performance metrics such as accuracy, precision, recall, and f1 score are used to assess the effectiveness of the algorithms in reducing the number of features needed for accurate classification.

The results in general demonstrate that the bio-inspired optimization algorithms are effective in reducing the number of features required for accurate classification. However, there have been variations in the performance of the algorithms on different datasets.
The study highlights the importance of data pre-processing and cleaning in ensuring the reliability and effectiveness of the analysis.

This study contributes to the advancement of predictive analytics in the realm of chronic diseases. The potential impact of this work extends to early intervention, precision medicine, and improved patient outcomes, providing new avenues for the delivery of healthcare services tailored to individual needs. The findings underscore the potential benefits of using bio-inspired optimization algorithms for feature selection in chronic disease prediction, offering valuable insights for improving healthcare outcomes.

\end{abstract}
%
%

\section{Introduction}
\label{sect:intro}
Chronic diseases represent a significant and escalating public health challenge, contributing to a substantial portion of global morbidity and mortality. Early detection and prediction of chronic diseases are vital for effective prevention, intervention, and personalized healthcare. In this era of data-driven decision-making, harnessing the power of advanced analytics and machine learning (ML) techniques holds the promise of revolutionizing early chronic disease prediction. Artificial intelligence techniques are used to predict diseases based on available patient data.

One of the most active areas in ML is supervised learning in different fields like cybersecurity \cite{ijhpcnLetteriPG19} \cite{cssLetteriPG18}, finance \cite{femibLetteriPGD23} \cite{ital-iaLetteri23} \cite{letteri2023volts}, formative assessment \cite{mis4telAngeloneLV23}, where a model is trained with a set of samples consisting of inputs and target outputs. Then, the trained model can predict the output of the samples that have never been seen before.
The dataset used by supervised learning is some sort of a matrix composed of samples (rows) and features (columns) that define the data. ML algorithms usually need large datasets. However, as the number of dimensions (features) increases, the amount of data needed to generalize the ML model accurately increases exponentially (the curse of dimensionality \cite{bellman1957dynamic}). To overcome the curse of dimensionality, we need to reduce the dataset finding a matrix similar to the original one but with fewer features. The new resulting dataset (matrix) can be used more efficiently than the original one.

Feature selection (FS) is one of the techniques used for dimensionality reduction \cite{LetteriCP21}. This technique consists of selecting the relevant features and discarding the irrelevant and redundant ones.
Reducing input dimension can lead to performance improvements by (i) decreasing the learning time and model complexity or; (ii) increasing the generalization capacity and accuracy. Furthermore, if suitable features are selected, this can result in a better understanding of the problem and reduce the measurement cost \cite{guyon2008feature}. 

FS has been successfully used in medical applications, in which the goal is, other than reducing the problem of dimensionality, reducing involved costs; such as the costs of extracting information from images \cite{remeseiro2013methodology}, and identifying the reasons for disagreements among experts about disease diagnosis \cite{bolon2015dealing}. In the medical field,
the analysis of illness is a challenging and tedious task. The description of the target class is not influenced by irrelevant characteristics. Unnecessary features create noise in classification and do not contribute to anything \cite{gultepe2019use}. They affect the classification results and make the system run slower. Thus, eliminating these features is essential before applying classifier algorithms. 

Bio-inspired optimization methods are an emerging area for solving complex real-world optimization problems. They are algorithms inspired by natural, biological, and ecological behaviour. They play an essential part in finding the best optimum solution in different problem domains. These algorithms show a high level of diversity, robustness, dynamic, simplicity, and interesting phenomena in comparison with other optimization methods. Recently, optimization algorithms have been used for supervised FS. Research interests in the use of bio-inspired methods for feature selection are growing due to their ability to deal with non-linear high-dimensional data.

 
This paper explores the novel paradigm of bio-inspired feature selection as a means to optimize predictive models for chronic diseases. Drawing inspiration from natural selection processes, we leverage Genetic Algorithms (GA), Particle Swarm Optimization (PSO), and Whale Optimization Algorithm (WOA) to identify a refined subset of features from complex and high-dimensional medical datasets. The core objective is twofold: to enhance the predictive accuracy of models and to streamline data dimensionality, ultimately making predictions more interpretable and actionable.

The research encompasses a comparative analysis of the three bio-inspired feature selection approaches mentioned above across diverse chronic diseases, including diabetes, cancer, kidney, and cardiovascular diseases. 



By bridging the gap between bio-inspired optimization techniques and machine learning, our research contributes to the advancement of predictive analytics in the realm of chronic diseases. The potential impact of this work extends to early intervention, precision medicine, and improved patient outcomes, providing new avenues for the delivery of healthcare services tailored to individual needs. In a world grappling with the burden of chronic diseases, the pursuit of innovative solutions to enhance prediction models is both timely and imperative.

The contribution of this work is:
\begin{itemize}
    \item A significant feature selection to eliminate redundant and inconsistent data through three bio-inspired optimization algorithms (GA, PSO, WOA), which improves the efficiency of the prediction model.
    \item To predict four chronic diseases like diabetes, breast cancer, heart, and kidney using Decision Tree, Random Forest, Neural Networks, Support Vector Machines, and Logistic Regression. The objective is early diagnosis of Chronic diseases.
\end{itemize}


This paper proceeds as follows: in Section \ref{sect:back}, we introduce feature selection and the three algorithms that we use in this study. Section \ref{sect:mat_meth} outlines the methodology adopted in this work. Section \ref{sect:results} presents the experiments and the practical findings achieved. In section \ref{sect:disc}, we carry out a discussion of the findings. Finally, in section \ref{sect:conclusion}, the study concludes by summarizing the main findings and limitations of this study, then outlining some possible future directions.

\section{Background}

\label{sect:back}
\subsection{Feature Selection Methods}
In this section, we focus on Feature Selection (FS) in the context of optimizing the prediction of chronic diseases. FS identifies the most relevant properties for accurate classification, reducing inputs to classifiers, and shortening training times. In the medical field, FS can pinpoint key factors for disease recognition, improving accuracy and reducing model complexity. The FS process is vital in various medical domains, including medical image processing, biomedical signal analysis, and DNA microarray data analysis. Most FS techniques in the medical field are filter-based, utilizing statistical or heuristic measures. However, FS is an NP-complete problem due to the vast space of possible combinations. For instance, with 10 features, there are 1024 combinations, underscoring the computational complexity of the problem \cite{dhal2022comprehensive}.

\subsubsection{Filter-Based Methods} assess feature relevance without the use of any learning algorithms, typically resulting in swift evaluations. In this model, features are appraised and categorized. However, these methods frequently overlook inter-feature dependencies, leading to similar features in the final set and the creation of weak and intricate learning models.
To tackle this issue, multivariate methods have been developed, considering feature dependencies in the ranking process.

\subsubsection{Wrapper-based models} use a classifier trained and employed to assess a set of relevant features. These methods employ an iterative search process in which each iteration of the learning model guides feature selection toward the optimal solution. However, due to the use of a learning model in the process, these methods often entail longer computation times and may sacrifice some degree of generality compared to filter-based methods, they also depend on a certain classification algorithm in evaluating the selected feature.

\subsubsection{Hybrid Methods} try to combine the main characteristics of filter and wrapper methods. In the filter phase, the feature dimensionality is reduced, and then the wrapper phase is used to select the most optimal feature subset. A filter-wrapper algorithm was proposed in \cite{guha2020embedded} which applied minimum redundancy maximum
relevance algorithm to carry out a local search mechanism. The authors in \cite{mafarja2019hybrid} used rough set theory and conditional entropy algorithm as filter methods for selecting the most relevant features from a set as the initial population. Then in the wrapper stage, they used a k-nearest neighbor (KNN) algorithm to evaluate the quality of the feature combination. The wrapper method determines the correlations in the features and thus can obtain a higher accuracy rate.

\subsubsection{Metaheuristic Optimization Methods} provide effective solutions to complex and NP-complete optimization challenges, delivering satisfactory results within reasonable time frames. Particularly valuable in managing high-dimensional datasets and mitigating computational complexities, these methods follow a fundamental process. 

Let $P$ be the initial population of solutions: $P = \{S_1, S_2, \dots, S_n\}$ aiming to enhance solution optimality through fitness evaluation at each iteration. The fitness of each solution is assessed through a fitness function, denoted as $F(S_i)$, with the cycle continuing until termination criteria are met.

Metaheuristic Optimization Methods are categorised into evolutionary-based algorithms (EAs) and swarm intelligence algorithms (SIAs), EAs draw inspiration from biological evolution, simulating Charles Darwin's "survival of the fittest" in the selection process. Genetic Algorithms (GA) exemplify this category \cite{gen1999genetic}. SIAs, emulating nature's collaboration, showcase self-organization and decentralization mechanisms observed in animals. Notable algorithms in this category include Particle Swarm Optimization (PSO) \cite{li2021improved} and Whale Optimization Algorithm (WOA) \cite{sharawi2017feature}.

In our study, we explore the bio-inspired optimization algorithms in feature selection across diverse medical datasets, offering valuable insights into their performance, considering these algorithms crucial in feature selection applications \cite{petwan2022review}.

\subsection{Genetic Algorithm}
\label{GA}
Genetic algorithms (GAs) are a class of process-inspired search and optimization algorithms of natural selection and the theory of evolution. The philosophy on which these algorithms are based is the survival of the fittest individual. GAs exploit the concept of ``natural selection'' to guide research towards better solutions over generations. The concepts for understanding a genetic algorithm are:

\subsubsection{Gene:} By \textit{Gene} we mean an indivisible unit of the algorithm that represents a subset of one single solution. In GAs, genes are usually represented as strings of bits or values, numbers, depending on the type of problem you are solving. In our case we will represent genes with Boolean values (0 or 1) and each gene will represent a single feature of the considered dataset.

\subsubsection{Population:} GAs are based on the evolution of the individuals contained in its population. By individual, we mean an abstract representation of the solution to the problem we want to address (solve), this solution is made up of a set of genes. In our case, we want to implement a GA that can evaluate which features must be selected from a dataset. We therefore need to represent the features in the dataset so that it is possible to evaluate which ones are combinations with the smallest number of features that impact performance as little as possible.
    
We will represent these solutions as arrays of genes where the value 1 is equivalent to using the feature and the value 0 equals waste. For example, the Pime Indian dataset has 9 columns of which eight represent the patients' biological values, such as glucose or blood pressure, and one represents the diagnosis. Since the diagnosis column is not a feature, as it represents the result we expect from the classifier, we have 8 features in the dataset. So we can represent the use of the first 3 features (glucose, blood pressure and skin thickness) as a vector of 8 elements (the number of elements is equal to the number of features) where the first three values to 1: [1, 1, 1, 0, 0, 0, 0, 0]. A population, or the set of various possible solutions to the problem, is made up of a set of individuals, where therefore each individual represents a solution. We therefore have the population be a two-dimensional array, where each row of the array represents an individual and, therefore a solution to the problem. In this case, the matrix 'population' has five solutions.
\begin{equation}
\begin{aligned}
\text{population} = 
\begin{bmatrix} 
    1 & 0 & 1 & 0 & 1 & 0 & 1 & 0 \\
    0 & 1 & 0 & 1 & 0 & 1 & 0 & 1 \\
    1 & 1 & 0 & 0 & 1 & 1 & 0 & 0 \\
    0 & 0 & 1 & 1 & 0 & 0 & 1 & 1 \\
    1 & 1 & 1 & 1 & 0 & 0 & 0 & 0 \\
\end{bmatrix}
\end{aligned}
\end{equation}
\label{matrice_binaria}
    
\subsubsection{Fitness Function:} The fitness function in a GA is a fundamental element that evaluates the goodness of the solutions represented by individuals within the population. Its definition is crucial as it guides the process of natural selection of the best solutions and, consequently, the evolution of the population.

The fitness function assigns a numerical value to each individual based on how well it represents an optimal solution for the problem at hand. This assessment is based on the specific objectives of the problem. For example, if we are trying to maximize or minimize a certain metric, the fitness function assigns a higher score to individuals who come closest to optimality according to a certain criterion, such as finding the shortest route between different cities.

The key to the success of a GA is that the fitness function must be defined appropriately so that better solutions receive a higher score than poorer-performing ones. This concept is also fundamental for the subsequent algorithms (\ref{PSO}) and (\ref{WOA}).

In our case, the objective is to reduce the number of features while maintaining accuracy in the classifications. It is therefore necessary to define a function that returns a greater value when, with the same accuracy, fewer features are used.

The approach proposed by \cite{li2021improved} suggests using the following formula:
    \begin{equation}
    \text{Fitness}(X) = \alpha \cdot \text{acc}(X) + (1 - \alpha) \cdot \frac{\#X}{N}
    \end{equation}
    Where:
    \begin{itemize}
        \item $\ alpha$ is a coefficient that weighs the accuracy against the cardinality of the set.
        \item $\text{acc}(X)$ represent the accuracy obtained with the features $X$.
    
        \item $\#X$ represent the cardinality (the number of elements) of the set $X$.
    
        \item $N$ represents the total number of features (i.e. the columns of the dataset, excluding the "diagnosis" one).
    \end{itemize}
    
    This formula combines the accuracy of the set $X$ with its cardinality, allowing us to evaluate which feature set is best, based on the goal of maximizing accuracy while maintaining a reasonable set size.
    
\subsubsection{Selection:} In the context of GA, the "selection" phase is one of the fundamental phases that simulates the process of natural selection. Its main function is to determine which individuals within a population should survive and be used to generate the next generation of individuals.
The selection is achieved, in our case, by taking the 4 individuals with the highest fitness.
These individuals will be saved for the next generation and new individuals will be created from them in the crossover and mutation phases.

\subsubsection{Crossover:} 
\begin{figure}[h]
  \centering
  \includegraphics[width=0.8\linewidth]{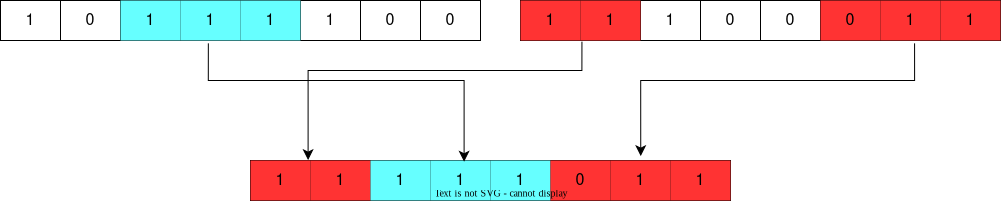}
  \caption{Example of crossover: at the top the parent solutions and at the bottom the solution created by crossover}
  \label{fig:crossover}
\end{figure}

The crossover phase allows the generation of new individuals starting from the "parents" chosen in the selection phase.
To perform the crossover, a random number of genes are taken from parent 1 and the rest of the individual is filled with those from the second parent.
Figure \ref{fig:crossover} shows how it was applied.

\subsubsection{Mutation:} 
\begin{figure}[h]
  \centering
  \includegraphics[width=0.8\linewidth]{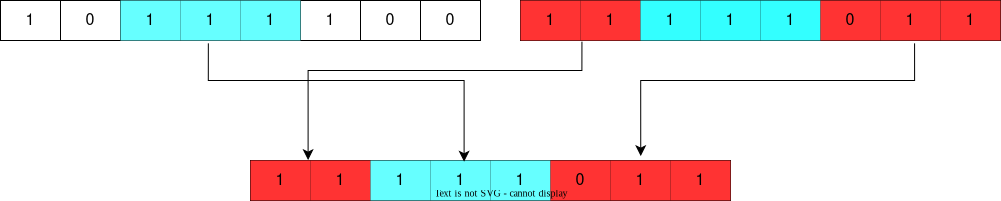}
  \caption{Example of crossover where a new individual is not generated.}
  \label{fig:perche_serve_mutation}
\end{figure}

In a GA, the "mutation" operation is fundamental and plays the role of introducing diversity into the population and promoting the exploration of the solution space. A mutation is responsible for the random modification of some characteristics or genes of individuals within the population. In this way, with each new generation, new unique individuals are created.
Although the crossover mixes the characteristics of the parents, this does not guarantee complete diversification, and some parts of the solution may remain unchanged (see figure \ref{fig:perche_serve_mutation}). 

Mutation introduces random variations even in those genes that were inherited from parents, thus improving diversity within the population.

\begin{figure}[h]
  \centering
  \includegraphics[width=0.6\linewidth]{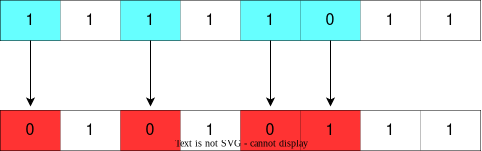}
  \caption{Mutation Example.}
  \label{fig:mutation}
\end{figure}

To implement the mutation, just make a random bit change from 0 to 1 and vice versa. In the implementation, $n/2$ genes are modified where n is the number of features figure \ref{fig:mutation}.

\begin{figure}[h]
  \centering
  \includegraphics[width=0.5\linewidth]{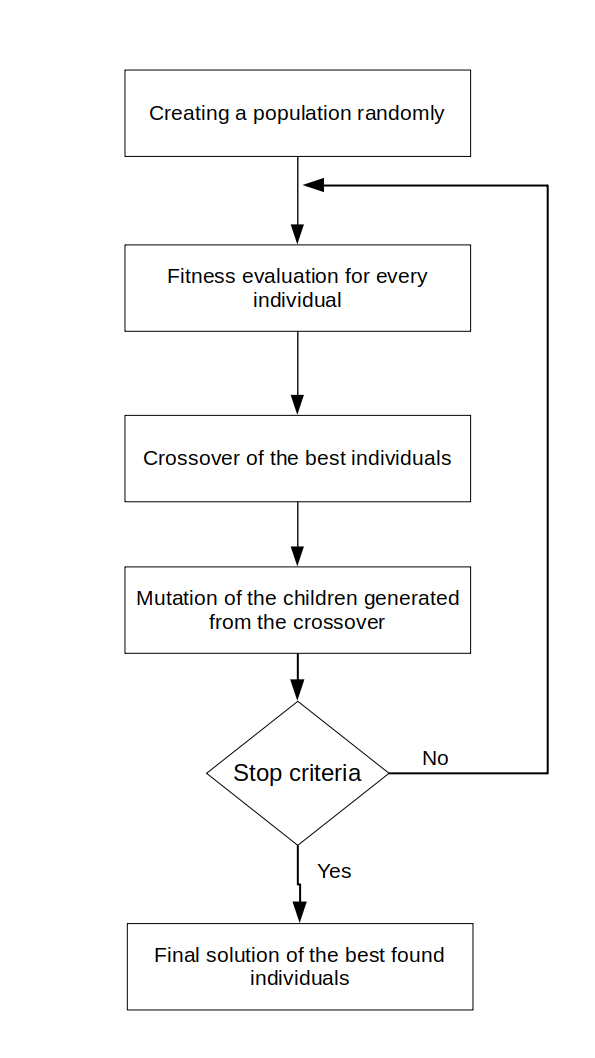}
  \caption{GA Operations Flow.}
  \label{fig:algo_gentico}
\end{figure}

For the one-generation cycle of a genetic algorithm, there are 4 fundamental steps: fitness evaluation, selection of the best individuals, crossover, and mutation figure \ref{fig:algo_gentico}.
These steps allow us to evolve an initial random population towards an optimal solution. It is not certain that the solution found is the best overall, but we will certainly arrive at a good solution with the right number of generations.
There are several loop termination criteria, for example:
\begin{itemize}
     \item Reaching a certain number of generations.
     \item The achievement of a fitness value.
     \item There has been no increase in fitness over the last $n$ generations.
\end{itemize}

\subsection{Particle Swarm Optimization}
\label{PSO}
Particle Swarm Optimization (PSO) is a population-based search and optimization technique inspired by the social behavior of birds moving in flocks. The main goal of PSO is to find the optimal solution for a given problem, where the “particle” represents a possible solution in the search space.

Particles
In this algorithm a solution is represented as a particle characterized by position eq. \ref{eq:position} and velocity eq. \ref{eq:velocity} that vary during iterations.
\begin{align}
X(t) &= (x(t), y(t)) \label{eq:position} \\
V(t) &= (Vx(t), Vy(t)) \label{eq:velocity}
\end{align}

At each iteration the particle position is updated according to the equation \ref{eq:position_update} and the velocity according to the \ref{eq:velocity_update} where $i$ represents the position of the particle and $t$ the i-th iteration .

\begin{equation}
X^i(t+1) = X^i(t) + V^i(t) \label{eq:position_update}
\end{equation}

\begin{equation}
    V^i(t+1) = w \cdot V^i(t) + c1 \cdot r1 \cdot (\text{pbest} - X^i(t) + c2\cdot r2 \cdot (\text{gbest} - X^i(t)) \label{eq:velocity_update}
\end{equation}

We can see that PSO has some parameters:
\begin{itemize}
    \item $w \in (0,1)$ represents inertia, it is a value that indicates how much the particle must maintain its current speed and direction.
    \item $c1$ is the cognitive coefficient and indicates how much the particle must concentrate on its best result. A high value of "c1" will cause a particle to take more into account the optimal solutions previously discovered by itself. 
    \item $c2$ is the social coefficient and indicates how much weight the particle gives to the results achieved by the group. A high value of "c2" will cause a particle to take more into account the optimal solutions discovered by other particles in the swarm.
    \item $r1$ e $r2 \in (0,1)$ they are random values that introduce randomness into the cognitive and social component of the particles. These values allow the particles to follow different trajectories and not get stuck in local optimal solutions.

    In figure \ref{fig:pso} we can see the search for the minimum of a function in the two-dimensional plane.
\begin{figure}[h]
  \centering
  \includegraphics[width=0.8\linewidth]{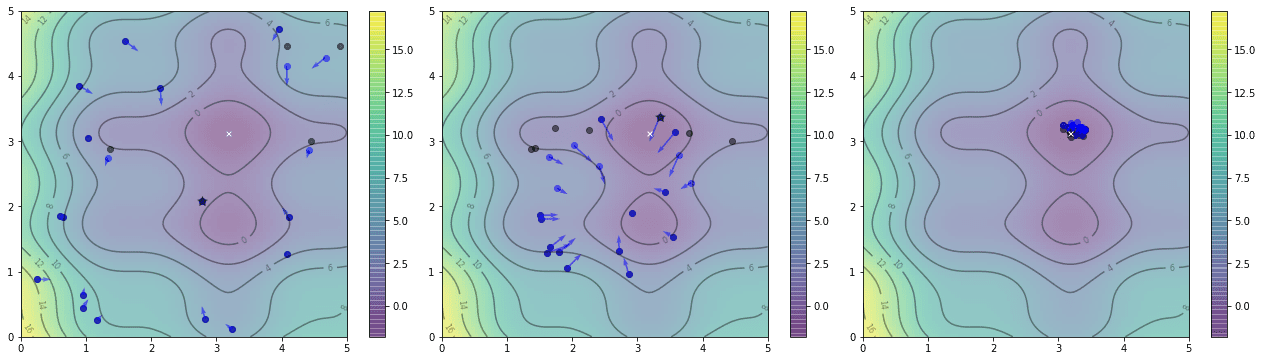}
  \caption{Finding the minimum of a function in a two-dimensional space (\url{https://shorturl.at/fglsu}).}
  \label{fig:pso}
\end{figure}
\end{itemize}

\subsubsection{Binary PSO.}
In our case, we are not working in a continuous space like the plane but we are looking for the solution to a problem represented as a binary vector. Kennedy and Eberhart's 1997 paper \cite{kennedy1997discrete} presented a discrete binary version of the PSO algorithm. The solution is to define the trajectories as the probability of the change of one bit of the solution vector from one state to another.

The formula for updating speeds expressed in equation \ref{eq:velocity_update} remain unchanged, while \ref{eq:position_update} is modified in the following way:
\begin{equation}
X^i(t+1) = 
\begin{cases}
   0 & \text{if} \text{ rand()} \geq S(V^i(t) \\
   1 & \text{if} \text{ rand()} < S(V^i(t)
\end{cases}
\label{sistema_pso}
\end{equation}

Where $S(x)$ is the sigmoid function to which we pass the velocity of the particle. This function allows us to map the set of values $(-\infty, +\infty)$ in the probability interval $(0,1)$.
\begin{equation}
    S(x) =  \frac{1}{1 + e^{-x}} \label{sigmoide}
\end{equation}


So just like for a genetic algorithm, we will have an initial population of particles represented as a random binary matrix where each row is a solution (\ref{matrice_binaria}).

The velocity of the particle is also limited in a range $[V_{min}, V_{max}] = [-4, 4]$ so that it does not explode due to the saturation velocity of the function making it impossible to change a bit during the search.

\subsubsection{Genotype-phenotype.}
In a more recent version by Lee et al. \cite{lee2008modified}, since in BPSO position update does not use position information, the concept of genotype and phenotype is suggested.
Genotype is the genetic basis of an individual, while phenotype is the visible or observable expression of those genes. For example, an individual's genotype may include the presence of genes for blue eyes, but the phenotype will be the actual appearance of the eyes.
The genotype-phenotype concept can be applied to the BPSO as follows: let the velocity and binary position of the original BPSO be a genotype $X_g$ and a phenotype $X_f$, respectively. Then, the equations \ref{eq:position_update}, \ref{eq:velocity_update} and \ref{sistema_pso} update functions of the original BPSO are modified as follows:

\begin{equation}
X_g^i(t+1) = X_g^i(t) + V^i(t) \label{eq:aggiornamento_posizione_2}
\end{equation}

\begin{equation}
    V^i(t+1) = w \cdot V^i(t) + c1 \cdot r1 \cdot (\text{pbest} - X_f^i(t) + c2\cdot r2 \cdot (\text{gbest} - X_f^i(t)) \label{eq:aggiornamento_velocità_nuovobpso}
\end{equation}

\begin{equation} \label{aggiornamento_pos_bpso}
X_f^i(t+1) = 
\begin{cases}
   0 & \text{if} \text{ rand()} \geq S(X_g^i(t) \\
   1 & \text{if} \text{ rand()} < S(X_g^i(t)
\end{cases}
\end{equation}

\subsubsection{Mutation.}
In \cite{lee2008modified} Lee et al. state that when a velocity converges near $V_{min}$ or $V_{max}$, it is difficult to change the corresponding position with a small change in velocity, which makes it difficult to escape from a local optimum.
A mutation probability $P_m$ is then inserted which acts as follows:
\begin{equation}
    \text{If } (rand() < P_m) \xrightarrow{} V(t+1) = -V(t+1)
\end{equation}
This velocity reversal will make it easier for particles not to get trapped in local optimum solutions.

The PSO algorithm seeks the best solution by simulating a flock where each individual/particle moves based on parameters. The search for a particle is influenced by the inertia value and by the cognitive and social coefficients.
In BPSO the particle speeds are mapped via a function into values between 0 and 1. These values will assign the bits of the solution vector the value 0 or 1 depending on the \ref{aggiornamento_pos_bpso}.

For each BPSO cycle, 4 basic steps occur for each particle: position update, genotype update, genotype mutation, and genotype decoding via the sigmoid transfer function.
As with a genetic algorithm, we have no guarantees that the algorithm will find the best possible solution, but a solution that comes close.
The loop termination criteria are the same as those stated in \ref{GA}.

\begin{figure}[h]
  \centering
  \includegraphics[width=0.5\linewidth]{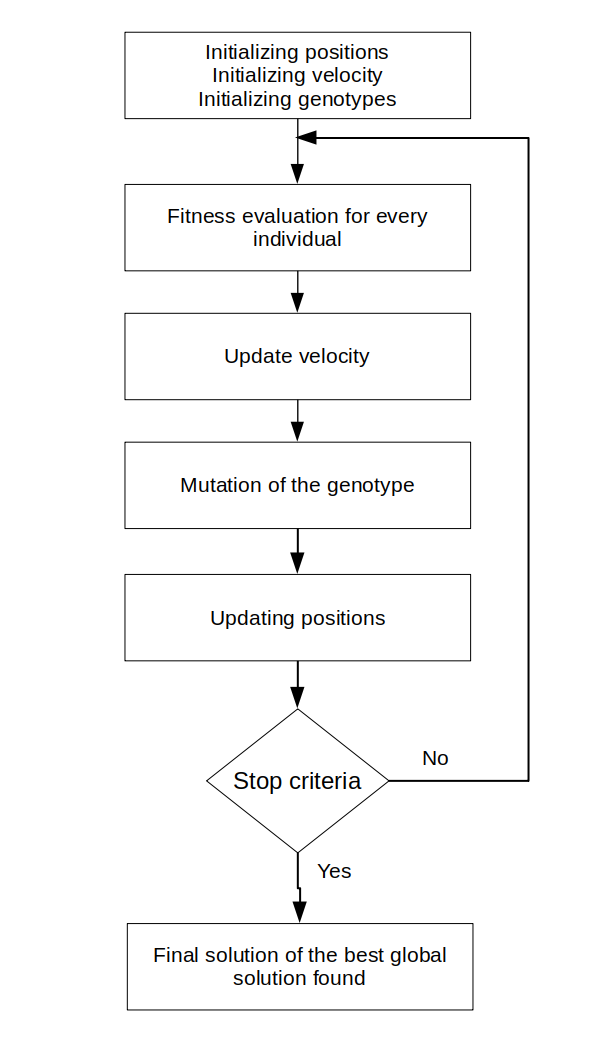}
  \caption{BPSO Operations flow.}
  \label{fig:bpso}
\end{figure}

\subsection{Whale Optimization Algorithm}
\label{WOA}
The Whale Optimization Algorithm (WOA) is a new search algorithm inspired by the hunting strategy of humpback whales \cite{sharawi2017feature}.
The strategy and mathematical formulation can be summarized as follows:

\paragraph{Prey encircling:} In this phase, the algorithm starts with an initial population and assumes that the current solutions are the best, these positions are the "prey". The other agents update their positions accordingly towards the best search agent. The best agent is chosen based on a fitness function.

\begin{align}
&\vec{D} = \left|\vec{C} \cdot \vec{X}^*(t) - \vec{X}(t)\right| \\
&\vec{X}(t + 1) = -\vec{X}^*(t) - \vec{A} \cdot \vec{D} \label{formula 1 woa}\\
&\vec{A} = 2\vec{a} \cdot \vec{r} - \vec{a}\\
&\vec{C} = 2 \cdot \vec{r}
\end{align}

where $t$ indicate the iteration, $\vec{A}$ e $\vec{C}$ are coefficients.
$\vec{X^*}$ is the position vector of the best position found and $\vec{X}$ is the position of the current agent.

$\vec{a}$ is linearly decremented from $2$ to $0$ during each iteration. It is therefore necessary to define a priori the number of iterations that you want the algorithm to perform.
The value $r$ is random $\in[0, 1]$ instead. This means that $\vec{A} \in [-a, a]$. But we will enter this phase only when $|\vec{A}| <$1 so that the agent gets closer to the best solution.

\paragraph{Exploitation phase:} This phase works through 2 approaches.
\begin{itemize}
    \item Shrinking encircling: In this phase the value of $\vec{a}$ is decreased from $2$ to $0$ during each iteration. This will lead to smaller and smaller movements around the best solution. When $a=0$ the agent will be in exactly the best position.

    \item Spiral updating: In this phase, the spiral movement of the whales around the prey is simulated.

    \begin{align}
        &\Vec{X(t+1)} = D' \cdot e^{bl} \cdot cos(2 \pi l) + \vec{X^{*}(t)} \label{formula 2 woa}\\
        &D' = |\vec{X^{*}(t)} - \vec{X(t)}|
    \end{align}

    We have $D'$ is the distance between $\vec{X^{*}(t)}$ and $\vec{X(t)}$, $b$ is a constant that determines the shape of the spiral and $l \in [0,1]$.
\end{itemize}

\paragraph{Exploration phase:} In this phase the agents search for prey based on their mutual positions if $|\vec{A}| > $1 the agents are forced to move away from the reference. Therefore this phase is opposed to the encircling part and pushes the agents towards exploration instead of towards the best solution. The mathematical model is similar to that of encircling:
\begin{align}
&\vec{D} = \left|\vec{C} \cdot \vec{X}_{rand}(t) - \vec{X}(t)\right| \label{formula 3 woa} \\
&\vec{X}(t + 1) = -\vec{X}_{rand}(t) - \vec{A} \cdot \vec{D} 
\end{align}

\begin{figure}[ht]
  \centering
  \includegraphics[width=0.5\linewidth]{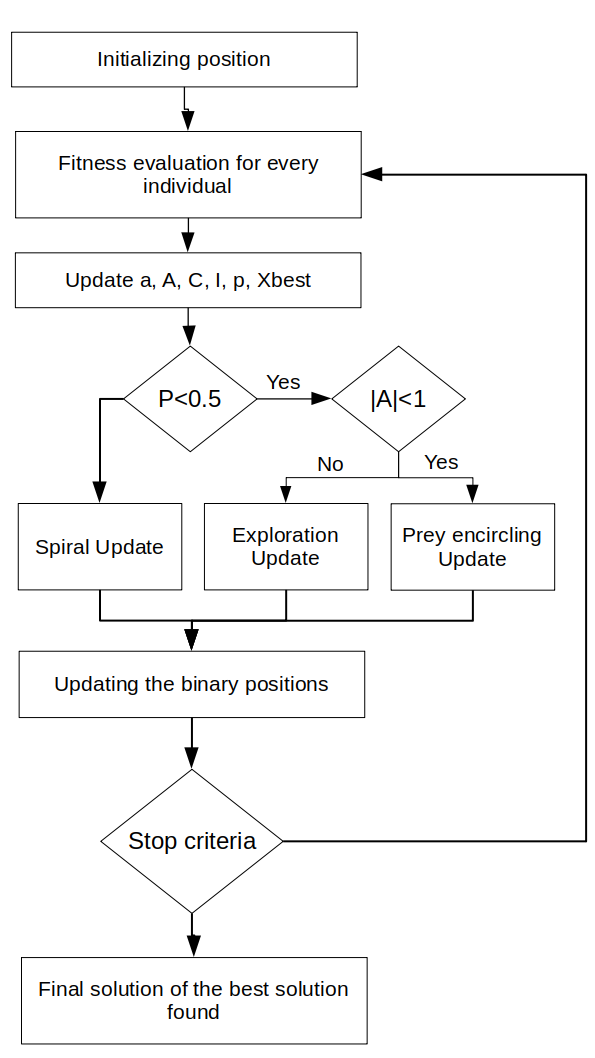}
  \caption{BWOA Operations Flow.}
  \label{fig:woa}
\end{figure}

The movement of the agents will then be based on applying the movements determined by \ref{formula 1 woa}, \ref{formula 2 woa} and \ref{formula 3 woa}. The algorithm is summarized in figure \ref{fig:woa}, and the cycle termination criteria are the same as those mentioned in Sec \ref{GA}.

\subsubsection{Binary WOA.}
Just like PSO the WOA algorithm is designed to move in a value space $\in \mathbf{R}$. We then need to map the real values into 0 and 1 values.
To do this we will also use the \ref{sigmoide} on each element of the vector of values representing the position of an agent.
Also for the BWOA the best individual is selected through the fitness function from Section \ref{GA}.

\section{Methodology}
\label{sect:mat_meth}
\begin{figure}[h]
  \centering
  \includegraphics[width=\linewidth]{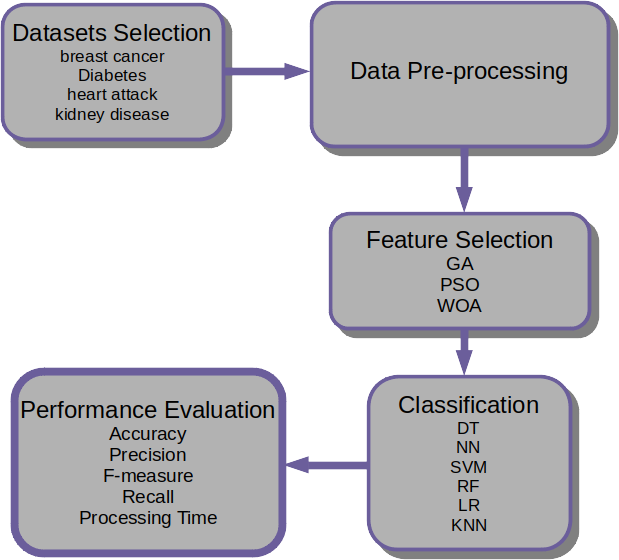}
  \caption{Methodology}
  \label{fig:method}
\end{figure}

Analysing chronic diseases like cancer, heart disease, diabetes, kidney disease, and many others is a significant challenge for doctors and medical practitioners. The main goal of this work is to increase the efficiency of the process of chronic disease prediction by using ML and FS techniques. Figure \ref{fig:method} shows the workflow of our methodology. The procedure of effective chronic disease prediction contains the following steps: (i) Data gathering, (ii) Data pre-processing, (iii) Feature Selection, (iv) Data classification, (v) Performance evaluation.

The datasets used in this work are obtained from different online data sources, one of the diabetes datasets is obtained from a hospital (the dataset was anonymised before using it). The pre-processing stage is very important because datasets can have errors, missing data, redundancies, noise, and many other problems that cause the data to be unsuitable to be used by the ML algorithm directly. This can be solved using data pre-processing techniques. It includes data transformation, data cleaning, missing values imputation, balancing, and data normalisation steps. FS techniques would be used to extract significant features. FS is performed using three algorithms viz. GA, PSO, and WOA. Data classification is performed using five ML algorithms viz. Decision Trees (DT), Random Forest (RF), Logistic Regression (LR), Neural Networks (NN), Support Vector Machines(SVM), and K-nearest Neighbour (KNN). Finally, the performance of different techniques is evaluated using different metrics.

\subsection{Problem Statement}
Medical data analysis is an essential and difficult task that needs to be carried out with extreme precision by considering the different factors. Chronic disease prediction is a repetitive task that needs great consideration to prevent misclassifications. A huge amount of data is usually required for this task. However, human interpretation of the data heavily relies on training and involvement, as a result, a computer-aided diagnosis system is an effective technique that can assist doctors in the early diagnosis of chronic diseases for more effective treatments and prevention. Medical datasets are commonly very large, and analytical precision is influenced by many variables, so it is extremely useful to reduce the dataset to those features that are best distinguished between the two cases or classes (e.g. normal vs. diseased) \cite{selvaraj2011microarray}. 
Chronic diseases are a growing health problem worldwide. Early diagnosis and prediction can enhance patients' survival rates by starting treatment early and changing their lifestyles. FS and particularly Bio-inspired FS and ML techniques permit the use of intellectual ways across different datasets for revealing valuable insights for decision-makers in medical diagnosis.

\subsection{Data Gathering}
\label{sect:mat}
We begin the discussion by describing the datasets used and providing a description of the origin, size, data types, and characteristics of each.
In each dataset, the columns represent the input parameters used to train the classifiers, while the rows indicate the amount of data sampled or recorded.
For each dataset there is a ``diagnosis" column with binary values that will be used for the supervised learning of the classifiers, the column has a value of "0" for a negative outcome and "1" for a positive outcome.

\subsubsection{Diabetes dataset.}
This dataset contains medical data collected from patients of the ASL-02 Lanciano-Vasto-Chieti. In the context of training the purpose of the classifiers will be to classify whether a patient is diabetic. This information is contained in the ``diagnosis" column which has a value of 0 to indicate a non-diabetic patient and 1 for diabetics.

General Information:
\begin{itemize}
    \item Number of rows:  2160
    \item Number of columns:  9
    \item Missing Data: 2535
    \item Features and missing data:

\begin{table}[H]
  \centering
  \begin{tabular}{lcc}
    \toprule
    \textbf{Variable} & \textbf{Data type} & \textbf{Number of missing data}\\
    \midrule
    sex                         & float64 & 0\\
    age                           & float64 & 0\\
    diagnosis                     & float64 & 0\\
    BMI value                    & float64 & 1134\\
    Blood sugar              & float64 & 213\\
    Creatinine            & float64 & 221\\
    Triglyceride          & float64 & 315\\
    Total Cholesterol    & float64 & 309\\
    HDL Cholesterol      & float64 & 343\\
    \bottomrule
  \end{tabular}
  \caption{List of Diabetes Dataset Features.}
  \label{tab:dati_diabete}
\end{table}

\end{itemize}

\subsubsection{Pime Indian dataset.}
This dataset comes from the National Institute of Diabetes and Digestive and Kidney Diseases. The goal of the dataset is to diagnostically predict whether or not a patient has diabetes. All patients here are women at least 21 years old of Pima Indian descent.

General Information:
\begin{itemize}
    \item Number of rows:  768
    \item Number of columns:  9
    \item Missing Data: 652
    \item Features and missing data:
 \begin{table}[H]
  \centering
  \begin{tabular}{lcc}
    \toprule
    \textbf{Variable} & \textbf{Datatype} & \textbf{Number of missing data} \\
    \midrule
    Glucose                     & float64 & 5\\
    BloodPressure               & float64 & 35\\
    SkinThickness               & float64 & 227\\
    Insulin                     & float64 & 374\\
    BMI                         & float64 & 11\\
    \multicolumn{2}{c}{\ldots} \\
   
    \bottomrule
  \end{tabular}
  \caption{List of variables Dataset Pime Indian.}
  \label{tab:dati_pime_indian}
\end{table}
\end{itemize}

\subsubsection{Breast Cancer dataset.}
The Wisconsin Breast Cancer dataset was created at the University of Wisconsin, Department of Computer Science, by William H. Wolberg. Data were collected from cytological examinations of aspirates with fine needle syringes (FNA) of breast masses. The objective of the classification is to recognize whether a tumor is benign or malignant. It can be found on Kaggle or UCI.

General Information:
\begin{itemize}
    \item Number of rows:  569
    \item Number of columns:  31
    \item Missing data: 0
    \item Features and missing data:
\begin{table}[H]
  \centering
  \label{tab:dati_bc}
  \begin{tabular}{lcc}
    \toprule
    \textbf{Variable} & \textbf{Data Type} & \textbf{Number of missing data} \\
    \midrule
    radius\_mean & float64 & 0\\
    texture\_mean & float64 & 0\\

    \multicolumn{2}{c}{\ldots} \\ 

    concave points\_worst & float64 & 0\\
    symmetry\_worst & float64 & 0\\
    fractal dimension\_worst & float64 & 0\\
    \bottomrule
  \end{tabular}
  \caption{List of Variables Breast Cancer Dataset.}
\end{table}
    \end{itemize}

\subsubsection{Kidney Disease.}
The data is medical information on chronic kidney disease, they were detected over 2 months in India. The dataset can be found on Kaggle or UCI.

General information:
\begin{itemize}
    \item Number of rows:  400
    \item Number of columns:  25
    \item Missing data: 1012
    \item Features and missing data:
\begin{table}[htbp]
  \centering
  \label{tab:dati_kd}
  \begin{tabular}{lcc}
    \toprule
    \textbf{Variable} & \textbf{Data type} & \textbf{Number of missing data} \\
    \midrule
     age          & float64&9 \\
    bp           & float64 &12\\
    sg           & float64 &47\\
    \multicolumn{2}{c}{\ldots} \\ 
    htn          & float64 &2\\
    dm           & float64 &2\\
    cad          & float64 &2\\
    \bottomrule
  \end{tabular}
  \caption{List of variables Dataset Kidney Disease.}
\end{table}
    \end{itemize}

\subsubsection{Heart failure dataset.}
This dataset contains the collected medical records of 299 heart failure patients during the follow-up period, where each patient profile presents 13 clinical characteristics. It can be found on kaggle.

General information:
\begin{itemize}
    \item Number of rows:  299
    \item Number of columns:  13
    \item Missing data: 0
    \item Features and missing data:
\begin{table}[H]
  \centering
  \label{tab:dati_hf}
  \begin{tabular}{llc}
    \toprule
    \textbf{Variable} & \textbf{Type of Data} & \textbf{Missed Data} \\
    \midrule
     age          & float64&0 \\
    anaemia           & float64 &0\\
    creatinine phosphokinase          & float64 &0\\
    \multicolumn{2}{c}{\ldots} \\ 
    time          & float64 &0\\
    smoking           & float64 &0\\
    sex          & float64 &0\\
    \bottomrule
  \end{tabular}
  \caption{List of Variables Dataset Heart Failure.}
\end{table}
    \end{itemize}

\subsection{Data Pre-processing}
The second step consists of data pre-processing in which we used different techniques for data pre-processing and cleaning for preparing the data before using them for training. This involved techniques for removing duplicates, identification and removal of anomalies, imputation for missing data, dataset balancing, and normalization.

\newpage

\subsubsection{Data Cleaning} aims to identify and rectify imperfections in raw data, including outliers and missing values, to ensure accurate analysis. This process is essential for maintaining data quality and reliability for subsequent analysis and modelling \cite{LetteriPVG20}. 

\begin{figure}{H}
\centering
  \includegraphics[width=0.6\linewidth]{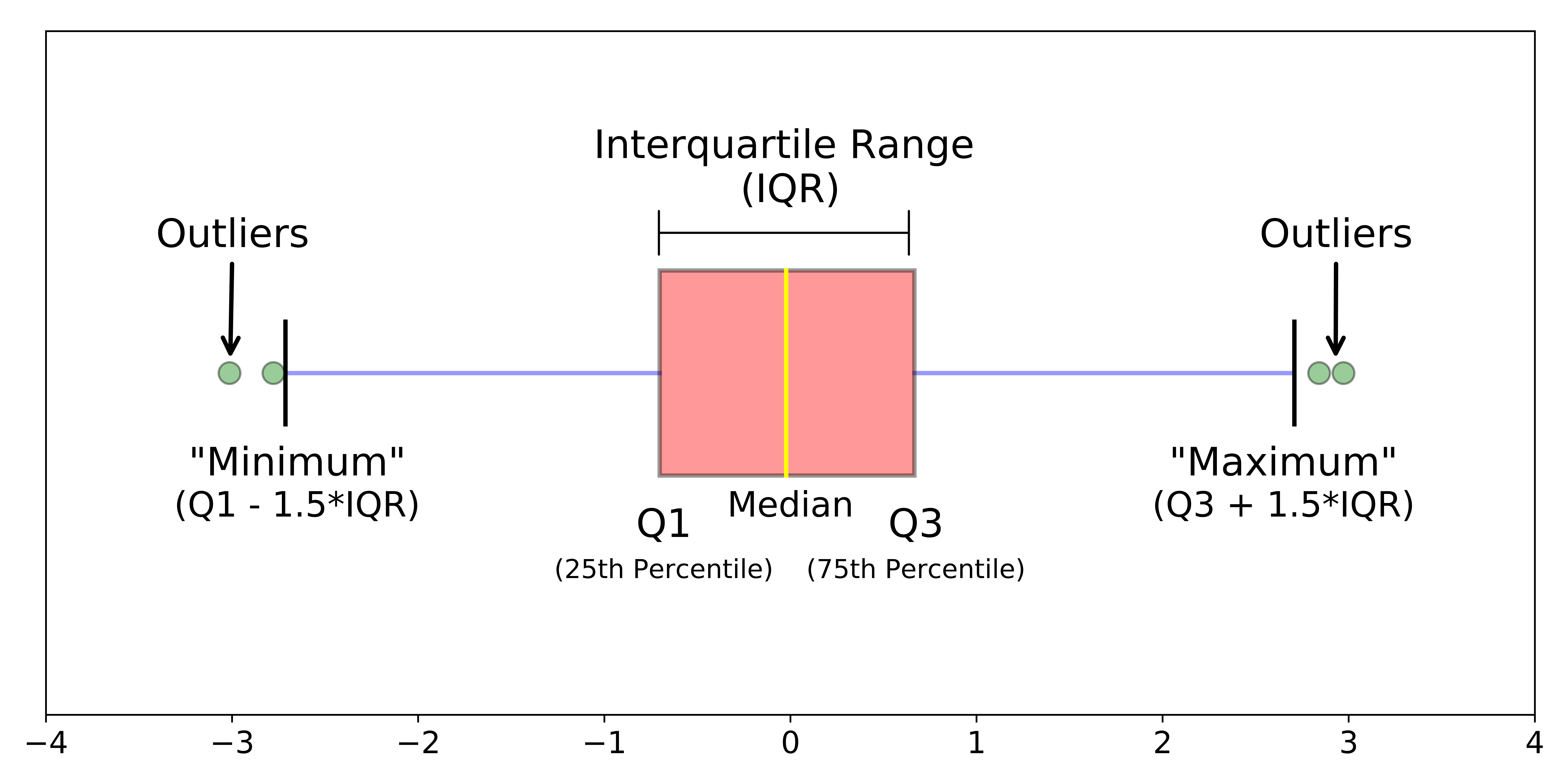}
  \centering
  \includegraphics[width=0.8\linewidth]{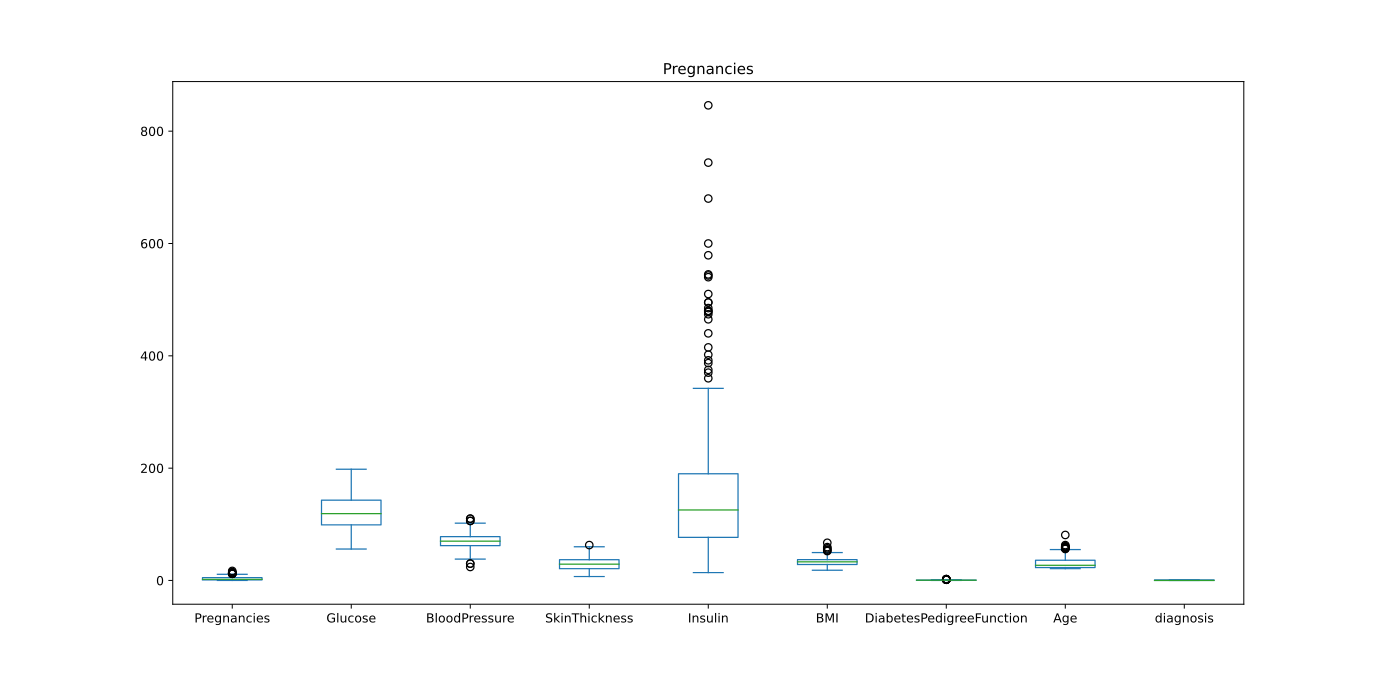}
  \caption{Boxplot of the dataset Pime Indian.}
  \label{fig:boxplot_pime_indian}
\end{figure}

Outliers, significantly deviating data points, are often addressed using the interquartile range (IQR) method, calculated as $IQR = Q3 - Q1$, where $Q3$ is the third quartile and $Q1$ is the first quartile. Figure \ref{fig:boxplot_pime_indian} shows an example of how the boxplot visually represents the IQR, median, and potential outliers in the case of Pime Indian dataset. 

\subsubsection{Missing Values Imputation} is the activity of addressing missing data, posing risks of performance degradation and biased results. Techniques involve deletion or imputation, where missing values are replaced. The K-Nearest Neighbors (KNN) algorithm excels in imputation, known for its adaptability to diverse data types. Given a dataset $D = \{(x_1,y_1),(x_2,y_2),\dots,(x_n,y_n)\}$, KNN imputes missing values $(\hat{x}_{missing})$ based on the average of $k$ nearest neighbors' feature values. The decision to implement KNN is supported by its demonstrated advantages in managing missing data and ensuring data integrity in machine learning. The imputation was performed with sklearn's KNNImputer with 5 neighbours.

\subsubsection{Data Balancing} is a critical concern due to the struggle of the classifiers when faced with disparate class distributions, leading to biased models. One popular approach to mitigate this issue is the Synthetic Minority Over-sampling Technique for Nominal and Continuous features (SMOTEEN) \cite{9505667}. This technique addresses imbalanced datasets by oversampling the minority class and, simultaneously, cleaning the majority class by combining the SMOTE (Synthetic Minority Over-sampling Technique) and Edited Nearest Neighbors (ENN) methods. The oversampling involves generating synthetic instances in the feature space, while ENN identifies and removes noisy examples. The combination of these processes enhances the model's ability to capture patterns in the minority class, fostering improved classification performance.

\subsubsection{Data Normalization} were applied and assessed on the distinct datasets using a neural network classifier. The min-max scaler exhibited marginally superior performance, prompting its adoption. This method scales dataset variables to a predefined range, typically between 0 and 1, though the range can be customized.

The Min-Max normalization formula, denoted by Equation (\ref{minmax_equation}), is employed.

\begin{equation}
\label{minmax_equation}
X_{\text{norm}} = \frac{X - X_{\text{min}}}{X_{\text{max}} - X_{\text{min}}}
\end{equation}

where $X_{norm}$ represents the normalized value of the feature, $X$ is the original value of the feature. $X_{min}$ denotes the minimum value among all feature values in the dataset, and $X_{max}$ denotes the maximum value among all feature values in the dataset.

This formula ensures that each feature is transformed proportionally to a specified range, enhancing the uniformity of data representation for improved machine learning outcomes.

\subsection{Bio-inspired Feature Selection}

After the data preparation phase, we applied the three bio-inspired feature selection algorithms (mentioned above) on each of the five datasets (see section \ref{sect:mat}).
Each algorithm uses the same fitness function described in (\ref{GA}), the value of $\alpha$ for the weighted sum has been set to 0.99 so that the main contribution of fitness is given to the accuracy of the classifications.
The classifier chosen for evaluating the fitness of the various solutions in the population of an algorithm is the K-Nearest Neighbors (KNN), this algorithm also used by \cite{sharawi2017feature} turns out to be much faster than classic machine learning algorithms as it does not require a real training phase. The number of neighbours used is 10.
The feature selection algorithms will have 20 agents as parameters and have 100 generations available.

\subsection{Performance Evaluations}
The first step in this phase is to divide the datasets into training and testing sets. We chose 70\% training and 30\% testing. The accuracy is calculated on 30\% of the testing dataset after training. The models used are:

\begin{itemize}
     \item Random Forest (RF) with 50 estimators
     \item Neural Network (NN) with 2 hidden layers of 10 neurons each.
     \item Decision Tree (DT)
     \item Supported Vector Machine (SVM)
     \item Logistic regression (LR)
     \item K-Nearest Neighbors (KNN) with 10 neighbours
\end{itemize}
For each dataset described in section \ref{sect:mat}, each model is trained and tested on 4 datasets: the WOA, PSO, GA features datasets and the dataset with all the features. This process is repeated 100 times for each dataset to have valid statistical data.
Furthermore, for each cycle, the division into train and test will be carried out again so that different portions of the dataset are used each time.

The confusion matrix, crucial in calculating performance metrics, illustrates a model's classification through the arrangement of $x_{tp}$ true positives, $x_{fp}$ false positives, $x_{fn}$ false negatives, and $x_{tn}$ true negatives.

The confusion matrix serves as the foundation for the following metrics: 
\begin{itemize}
    \item \textit{Accuracy} measures the proportion of correct classifications in a model as follows: $\frac{x_{tp}+x_{tn}}{x_{tp}+x_{fp}+x_{fn}+x_{tn}}$.
    \item \textit{Recall} is used to measure a model's ability to find positives as follows: $\frac{x_{tp}}{x_{tp}+x_{fn}}$.
    \item \textit{Precision} is used to measure the accuracy of positive predictions as follows: $\frac{x_{tp}}{x_{tp}+x_{fp}}$.
    \item \textit{F1-score} is used to measure a model's precision and recall balance as follows: $2 \times \frac{Precison \times Recall}{Precison + Recall}$.
\end{itemize}

\section{Experiments and Results}
\label{sect:results}

\subsection{Breast Cancer Dataset Experiments and Reults:}

\begin{figure}[h]
  \centering
  \includegraphics[width=0.6\linewidth]{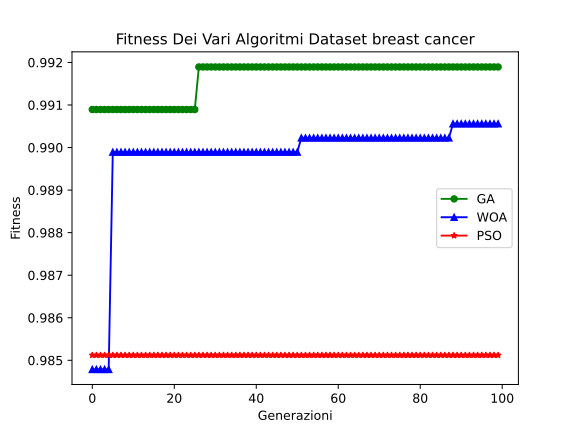}
  \caption{Performance of Genetic Algorithm, Particle Swarm Optimization and Whale Optimization Algorithm on the Breast Cancer dataset.}
  \label{fig:fitness_bc}
\end{figure}

Breast Cancer dataset after the preprocessing phase has:
\begin{itemize}
    \item Number of rows: 609
    \item Number of features: 30
\end{itemize}
Figure \ref{fig:fitness_bc} shows the fitness trend of the various algorithms over the 100 generations.

The final features selected by the various algorithms are:
\begin{itemize}
    \item WOA Features:  ['smoothness mean', 'symmetry mean', 'perimeter se', 'area se', 'smoothness se', 'concavity se', 'perimeter worst', 'area worst', 'smoothness worst', 'concavity worst', 'symmetry worst', 'fractal dimension worst']
    
\item PSO Features:  ['radius mean', 'area mean', 'smoothness mean', 'compactness mean', 'fractal dimension mean', 'radius se', 'texture se', 'area se', 'smoothness se', 'compactness se', 'texture worst', 'concavity worst']

\item GA feature:  ['texture mean', 'concavity mean', 'area se', 'compactness se', 'concave points se', 'fractal dimension se', 'radius worst', 'compactness worst']
\end{itemize}

The table \ref{tab:riassunto_fitness_bc} summarize the obtained performance
\begin{table}[H]
  \centering
  \begin{tabular}{llll}
    \toprule
    \textbf{Algorithm} & \textbf{Fitness}  &\textbf{\#Features} & \textbf{Reduction}\\
    \midrule
    GA  & $\approx$ 0.992 & 8 & 73.3\% \\
    WOA & $\approx$ 0.9905 & 12 & 60\% \\
    PSO & $\approx$ 0.985 & 12 & 60\% \\
    \bottomrule
  \end{tabular}
  \caption{Performance in Dataset Dimension Reduction.}
  \label{tab:riassunto_fitness_bc}
\end{table}

\begin{figure}[h]
  \centering
  \includegraphics[width=1\linewidth]{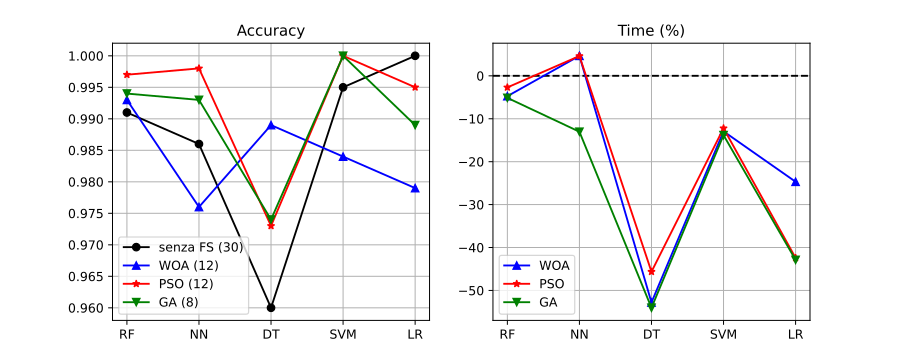}
  \caption{Accuracy and Time obtained on Breast Cancer dataset.}
  \label{fig:risultati_fs_bc}
\end{figure}

Now that we have the features selected we can test the performance of various algorithms on the dataset containing only the GA, PSO and WOA features and the one with all the features,
figure \ref{fig:risultati_fs_bc} shows in the graph on the left the accuracies of the various classifiers with the various features and on the right the percentage variation of the training time before and after the FS.
For optimal results we expect accuracy to remain almost unchanged and training times to decrease. We notice a decrease in training time when the points are below the value 0 on the Y-axis.

Table \ref{tab:riassunto_prestazioni_bc} summarizes the obtained performance results with various algorithms.
\begin{table}[H]
  \centering
  \begin{tabular}{llllll}
    \toprule
      &\textbf{RF} & \textbf{NN}  &\textbf{DT} & \textbf{SVM} &\textbf{LR}\\
    \midrule
    No FS  &  99.1\%        & 98.6\%           & 96\%           & 99.5\% & 100\% \\
    WOA &  \textbf{99.3\%} & 97.6\% & \textbf{98.8\%} & 98.4\% & 97.9\% \\
    PSO &  \textbf{99.6\%} & \textbf{99.7\%} & \textbf{97.4\%} & \textbf{100\%} & 99.5\%\\
    GA &  \textbf{99.4\%} & \textbf{99.4\%} & \textbf{97.4\%} & \textbf{100\%} & 98.9\%\\
    \bottomrule
  \end{tabular}
  \caption{Accuracy}
  \label{tab:riassunto_prestazioni_bc}
\end{table}

\begin{figure}[h]
  \centering
  \includegraphics[width=1\linewidth]{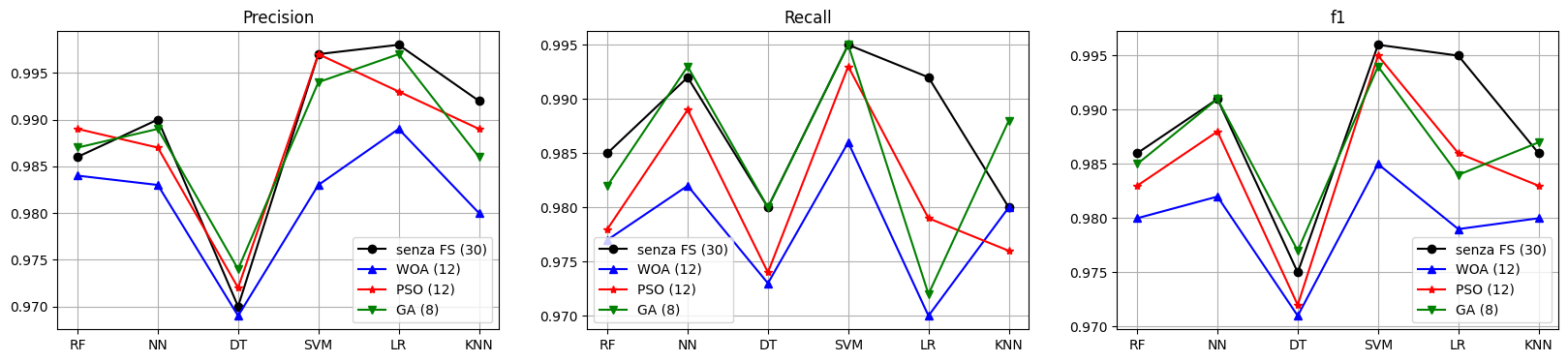}
  \caption{Precision, Recall and F1-score on dataset Breast Cancer.}
  \label{fig:acc_recall_f1_bc}
\end{figure}

Regarding the training time, we can observe in the right graph in figure \ref{fig:risultati_fs_bc} that the training times have globally decreased up to a maximum of 53\% for WOA and GA in the case of Decision Tree.
For the neural network in some cases (with PSO and WOA) the time has increased. This is due to an increase in the number of training epochs of the model to reach convergence. Decreasing the features may have made the data more difficult for the neural network to generalize. Further metrics are shown in figure \ref{fig:acc_recall_f1_bc}. 

\subsection{Experiments and Results with Pime Indian Dataset} 
\begin{figure}[h]
  \centering
  \includegraphics[width=0.6\linewidth]{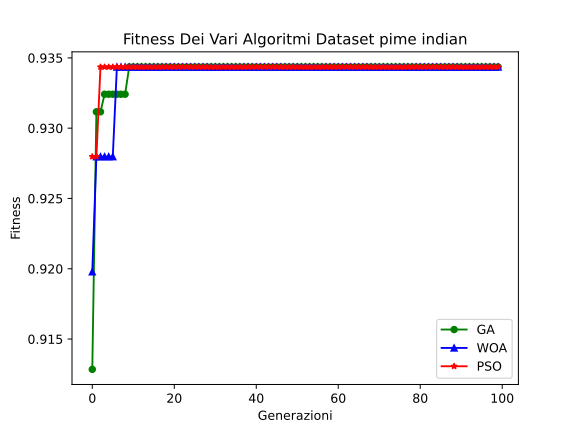}
  \caption{Performance of Genetic Algorithm, Particle Swarm Optimization and Whale Optimization Algorithm on the Pime Indian dataset.}
  \label{fig:fitness_pi}
\end{figure}

The dataset Pime Indian after the pre-processing phase has:
\begin{itemize}
    \item Number of rows: 559
    \item Number of features: 8
\end{itemize}
Figure \ref{fig:fitness_pi} shows the fitness trends of the various algorithms over 100 generations.

The final features selected by the various algorithms are:
\begin{itemize}
    \item WOA Feature:  ['Pregnancies', 'Glucose', 'BloodPressure', 'SkinThickness', 'Insulin', 'BMI', 'Age'].
    \item PSO Feature:  ['Pregnancies', 'Glucose', 'BloodPressure', 'SkinThickness', 'Insulin', 'BMI', 'Age']
    \item GA feature:  ['Pregnancies', 'Glucose', 'BloodPressure', 'SkinThickness', 'Insulin', 'BMI', 'Age']
\end{itemize}

we can see that in this case, all the algorithms agree on the features to be selected they all achieved the same fitness value. Table \ref{tab:riassunto_fitness_bc} summarises the performance.
\begin{table}[htbp]
  \centering
  \begin{tabular}{llll}
    \toprule
    \textbf{Algorithm} & \textbf{Fitness}  &\textbf{\#Features} & \textbf{Reduction}\\
    \midrule
    GA, PSO, WOA  & $\approx$ 0.934 & 7 & 12.5\% \\
    \bottomrule
  \end{tabular}
  \caption{Performance in feature reduction.}
  \label{tab:riassunto_fitness_pi}
\end{table}

The following results were obtained in the test:
\begin{figure}[H]
  \centering
  \includegraphics[width=1\linewidth]{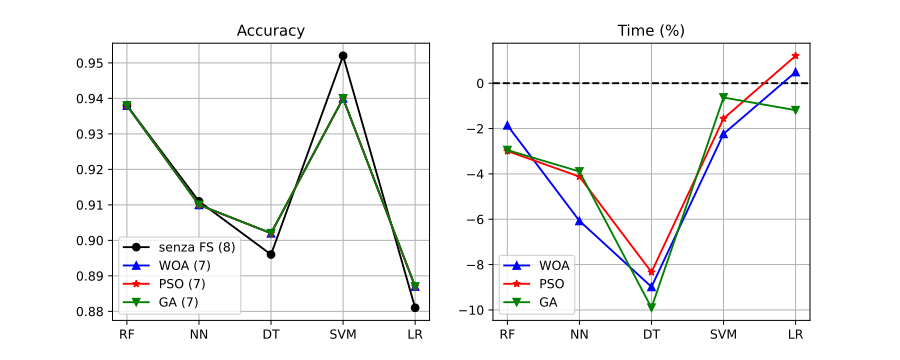}
  \caption{Accuracy and Time on dataset Pime Indian.}
  \label{fig:risultati_fs_pi}
\end{figure}

\newpage

The table \ref{tab:riassunto_prestazioni_pi} summarizes the accuracy obtained.
\begin{table}[H]
  \centering
  \begin{tabular}{llllll}
    \toprule
      &\textbf{RF} & \textbf{NN}  &\textbf{DT} & \textbf{SVM} &\textbf{LR}\\
    \midrule
    No FS  &  93.9\%        & 91\%           & 89.9\%           & 95\% & 88.1\% \\
    FS &  93.9\% & 91\% & \textbf{90.1\%} & 94\% & \textbf{88.8\%} \\
    \bottomrule
  \end{tabular}
  \caption{Accuracy Results.}
  \label{tab:riassunto_prestazioni_pi}
\end{table}

\begin{figure}[h]
  \centering
  \includegraphics[width=1\linewidth]{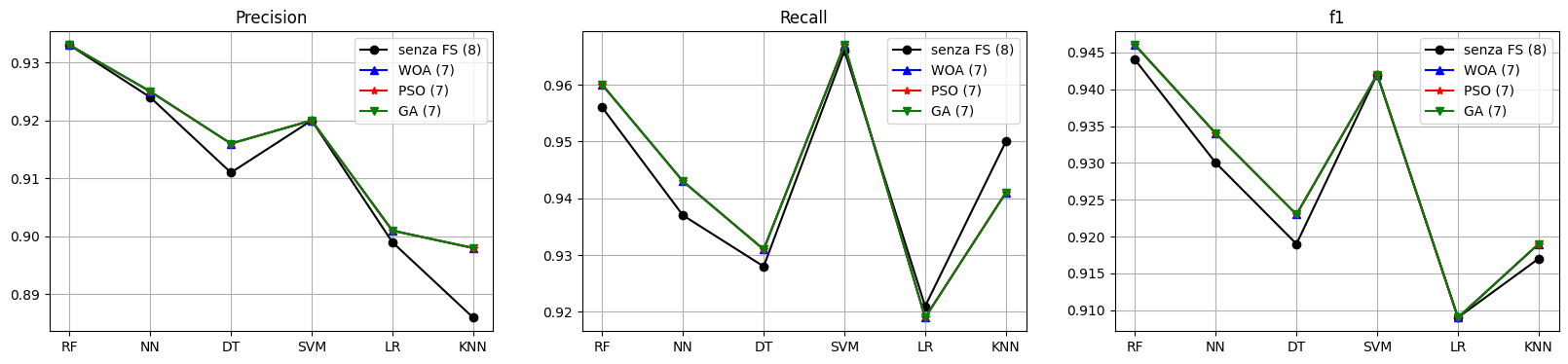}
  \caption{Precision, Recall and F1-score on dataset Pime Indian.}
  \label{fig:acc_recall_f1_pi}
\end{figure}

In this dataset, the global results are very similar to those without FS, in fact only one feature has been removed. For accuracy, there was a maximum decrease of 1\%.
Training times decreased by a maximum of 10\%. with small variations between the various algorithms. Since there is a deviation of $\approx 2\%$ between GA and PSO time in the LR despite there being no variations in the settings of the model parameters or features used. Further metrics are shown in figure \ref{fig:acc_recall_f1_pi}. 

\subsection{Experiments and Results with Heart Failure Dataset}
\begin{figure}[h]
  \centering
  \includegraphics[width=0.6\linewidth]{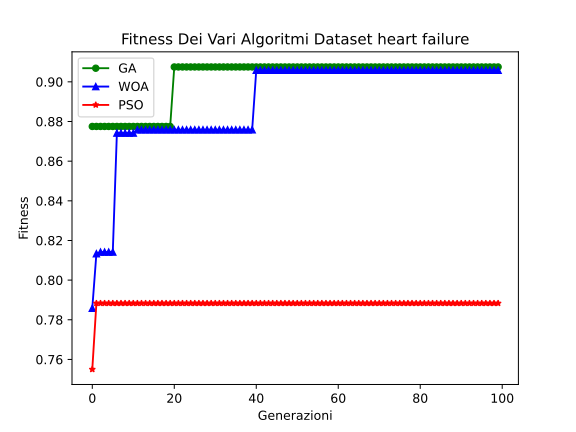}
  \caption{Performance of Genetic Algorithm, Particle Swarm Optimization and Whale Optimization Algorithm on the Heart Failure dataset.}
  \label{fig:fitness_hf}
\end{figure}

The dataset Breast Cancer after the preprocessing phase has:
\begin{itemize}
    \item Number of rows: 103
    \item Number of features: 12
\end{itemize}
Figure \ref{fig:fitness_hf} shows the fitness trend of the various algorithms over the 100 generations.

The final features selected by the various algorithms are:
\begin{itemize}
    \item  WOA Features:  ['age', 'anaemia', 'diabetes', 'serum creatinine', 'serum sodium']
    \item PSO Features:  ['platelets', 'serum creatinine']
    \item GA feature:  ['platelets', 'serum sodium', 'time']
\end{itemize}

Table \ref{tab:riassunto_fitness_hf} summarizes the performance obtained.
\begin{table}[htbp]
  \centering
  \begin{tabular}{llll}
    \toprule
    \textbf{Algoritm} & \textbf{Fitness}  &\textbf{\#Features} & \textbf{Reduction}\\
    \midrule
    GA  & $\approx$ 0.91 & 3 & 75\% \\
    WOA & $\approx$ 0.91 & 5 & 58\% \\
    PSO & $\approx$ 0.79 & 2 & 83.3\% \\
    \bottomrule
  \end{tabular}
  \caption{Performance in features reduction.}
  \label{tab:riassunto_fitness_hf}
\end{table}

We note that GA and WOA obtained the same fitness with a different number of features.
The following results were obtained in the test:

\begin{figure}[h]
  \centering
  \includegraphics[width=1\linewidth]{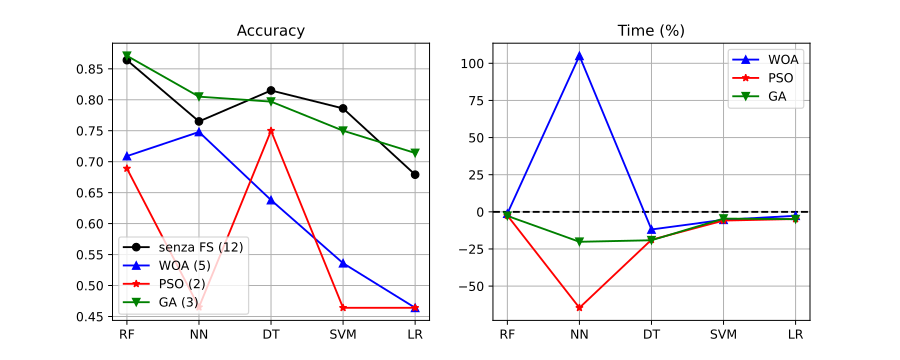}
  \caption{Accuracy and Time dataset Heart Failure.}
  \label{fig:risultati_fs_hf}
\end{figure}

The table \ref{tab:riassunto_prestazioni_hf} summarizes the accuracy and time obtained.

\begin{table}[H]
  \centering
  \begin{tabular}{llllll}
    \toprule
      &\textbf{RF} & \textbf{NN}  &\textbf{DT} & \textbf{SVM} &\textbf{LR}\\
    \midrule
    No FS  &  86\%        &76\%           & 81\%           & 78\% & 68\% \\
    WOA &  71\% & 75\% & 64\% & 54\% & 46\% \\
    PSO &  69\% & 46\% & 75\% & 46\% & 46\%\\
    GA &  \textbf{87\%} & \textbf{81\%} & 80\% & 75\% & \textbf{71\%}\\
    \bottomrule
  \end{tabular}
  \caption{Accuracy}
  \label{tab:riassunto_prestazioni_hf}
\end{table}

\begin{figure}[h]
  \centering
  \includegraphics[width=1\linewidth]{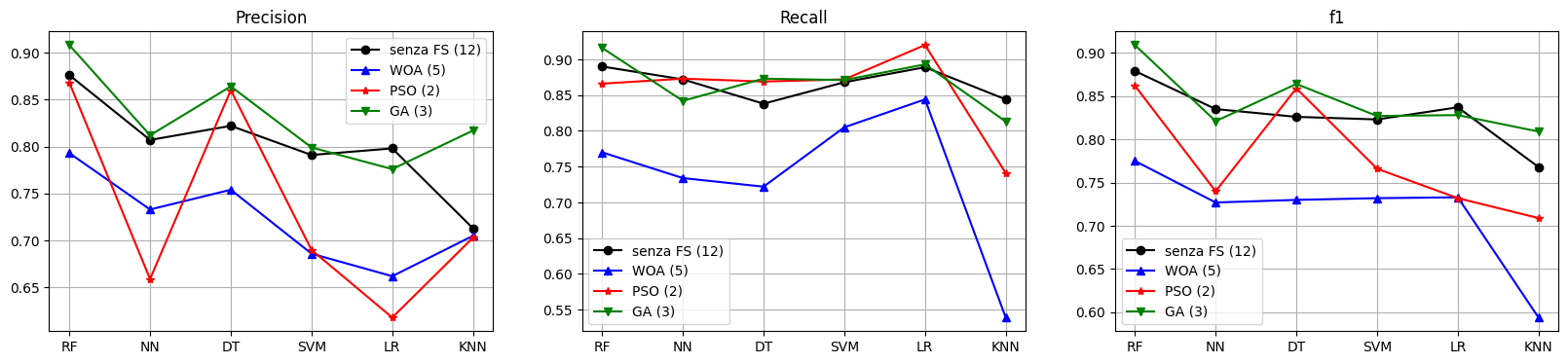}
  \caption{Precision, Recall and F1-score on dataset Heart Failure.}
  \label{fig:acc_recall_f1_hf}
\end{figure}

In this dataset, the overall performance was worse for all algorithms except the genetic algorithm. This is probably due to the low amount of data available which did not allow the model to have enough data for learning.
The genetic algorithm significantly outperformed WOA even though the calculated fitness for both was the same.
This is because the fitness was calculated on a different model with a different generalization ability than the others.
In this dataset, the training time of the neural network is even doubled. This is always due to a larger number of iterations needed for convergence as the data is decreased.
In general, however, the training times have all decreased, with the GA features the time has decreased by 24\% in the most relevant case of the neural network.
The genetic algorithm achieved excellent results, in fact with a 75\% reduction in the number of features it managed to obtain greater accuracy than using the entire set in 3 out of 4 models. Increasing the accuracy by 5\% in the most relevant case. Further metrics are shown in figure \ref{fig:acc_recall_f1_hf}. 

\subsection{Experiments and Results on Kidney Disease Dataset}
\begin{figure}[h]
  \centering
  \includegraphics[width=0.6\linewidth]{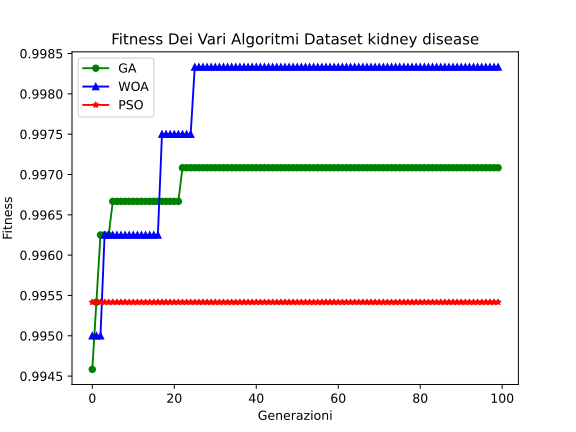}
  \caption{Performance of Genetic Algorithm, Particle Swarm Optimization and Whale Optimization Algorithm on the Kidney Disease dataset.}
  \label{fig:fitness_kd}
\end{figure}

The dataset Kidney Disease after the pre-processing phase has:
\begin{itemize}
    \item Number of rows: 238
    \item Number of features: 24
\end{itemize}
Figure \ref{fig:fitness_kd} shows the fitness trend of the various algorithms over the 100 generations.

The final features selected by the various algorithms are:
\begin{itemize}
    \item WOA Features:  ['rbc', 'bgr', 'bu', 'pcv']
    \item PSO Features:  ['age', 'su', 'rbc', 'pc', 'bgr', 'sod', 'pot', 'hemo', 'pcv', 'rc', 'cad']
    \item GA feature:  ['rbc', 'bgr', 'sod', 'hemo', 'pcv', 'dm', 'cad']
\end{itemize}

The table \ref{tab:riassunto_fitness_kd} summarizes the performance obtained.
\begin{table}[htbp]
  \centering
  \begin{tabular}{llll}
    \toprule
    \textbf{Algoritm} & \textbf{Fitness}  &\textbf{\#Features} & \textbf{Reduction}\\
    \midrule
    GA  & $\approx$ 0.998 & 7 & 70\% \\
    WOA & $\approx$ 0.997 & 4 & 83.3\% \\
    PSO & $\approx$ 0.995 & 11 & 54\% \\
    \bottomrule
  \end{tabular}
  \caption{Performance in features reduction.}
  \label{tab:riassunto_fitness_kd}
\end{table}

The following results were obtained in the test:
\begin{figure}[H]
  \centering
  \includegraphics[width=0.9\linewidth]{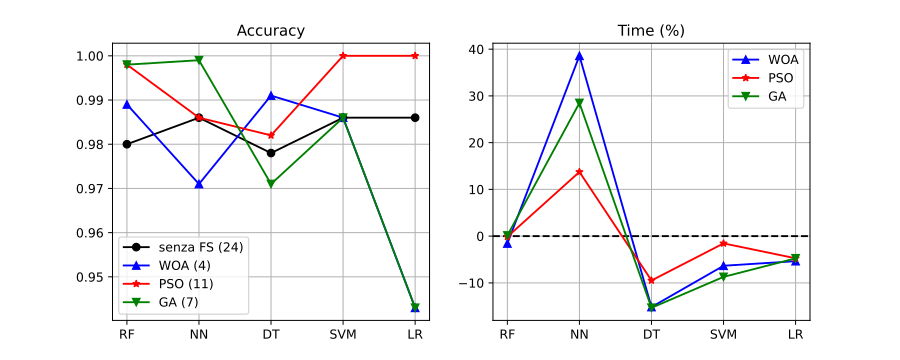}
  \caption{Accuracy and Time on dataset Kidney Disease.}
  \label{fig:risultati_fs_kd}
\end{figure}

The table \ref{tab:riassunto_prestazioni_kd} summarize accuracy obtained:

\begin{table}[H]
  \centering
  \begin{tabular}{llllll}
    \toprule
      &\textbf{RF} & \textbf{NN}  &\textbf{DT} & \textbf{SVM} &\textbf{LR}\\
    \midrule
    No FS  &  98\%        &98.8\%           &97.9\%           & 98.6\% & 98.6\% \\
    WOA &   98\% & 97.1\% & \textbf{99.1\%} & 98.6\% & 94.6\%\\
    PSO & \textbf{99.8\%} & 97.1\% & 98.1\% & \textbf{100\%} & \textbf{100\%}\\
    GA &  \textbf{99.8\%} & \textbf{99.9\%} & 97.1\% & 98.6\% & 94.6\%\\
    \bottomrule
  \end{tabular}
  \caption{Accuracy}
  \label{tab:riassunto_prestazioni_kd}
\end{table}

\begin{figure}[h]
  \centering
  \includegraphics[width=1\linewidth]{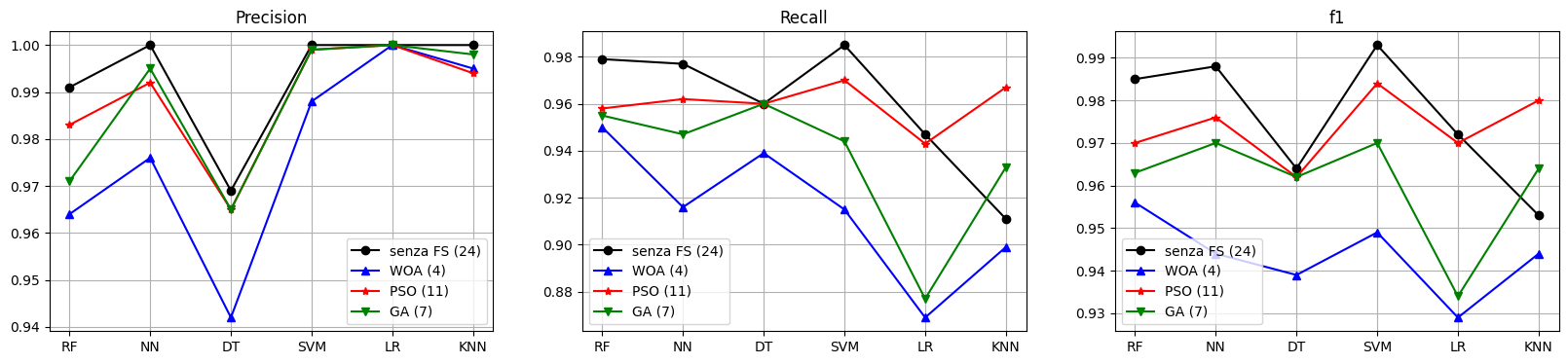}
  \caption{Precision, Recall and F1-score on dataset Kidney Disease.}
  \label{fig:acc_recall_f1_kd}
\end{figure}

In this dataset, high performance was achieved with most models. Although the GA was the algorithm with the highest fitness, it achieved lower performance than the PSO, which was the one with slightly lower fitness of all, in 3 out of 5 models. The PSO even achieved 100\% accuracy in two datasets exceeding the accuracy with the full set of features by 1.4\%.
The training time here has also decreased globally except for the neural network. Maximum improvements of 15\% of training time were achieved with the GA and WOA feature sets on the SVM model. Further metrics are shown in figure \ref{fig:acc_recall_f1_kd}. 

\subsection{Experiments and Results on Diabete Dataset}
 The diabetes dataset after the pre-processing phase has:
\begin{itemize}
    \item Number of rows: 1763
    \item Number of features: 8
\end{itemize}

\begin{figure}[h]
  \centering
  \includegraphics[width=0.6\linewidth]{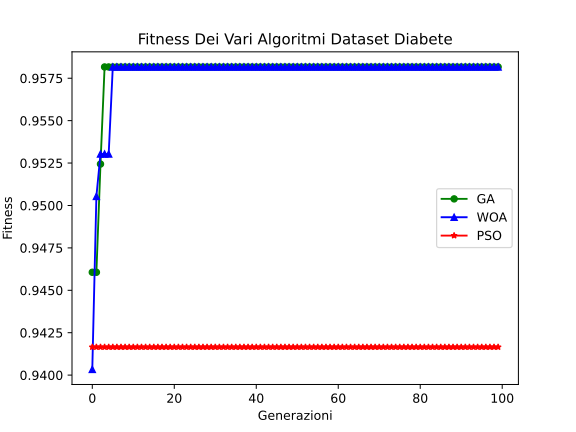}
  \caption{Performance of Genetic Algorithm, Particle Swarm Optimization and Whale Optimization Algorithm on the Diabetes dataset.}
  \label{fig:fitness_d}
\end{figure}

Figure \ref{fig:fitness_d} shows the fitness trend of the various algorithms over the 100 generations.

The final features selected by the various algorithms are:
\begin{itemize}
    \item WOA Features:  ['sex', 'age', 'BMI value', 'Blood Sugar', 'Creatinine', 'Total Cholesterol']
    \item PSO Features:  ['sex', 'eta', 'BMI Value', 'Blood Sugar', 'Creatinine', 'Total Cholesterol', 'Cholesterol Hdl']
    \item GA Features:  ['sex', 'age', 'BMI value', 'Blood Sugar', 'Creatinine', 'Total Cholesterol']
\end{itemize}

The table \ref{tab:riassunto_fitness_d} summarizes the performance results.
\begin{table}[htbp]
  \centering
  \begin{tabular}{llll}
    \toprule
    \textbf{Algorithm} & \textbf{Fitness}  &\textbf{\#Features} & \textbf{Reduction}\\
    \midrule
    GA, WOA  & $\approx$ 0.958 & 6 & 14.3\% \\
    PSO & $\approx$ 0.943 & 7 & 12.5\% \\
    \bottomrule
  \end{tabular}
  \caption{Performance in features reduction.}
  \label{tab:riassunto_fitness_d}
\end{table}

The following results were obtained in the test:

\begin{figure}[h]
  \centering
  \includegraphics[width=1\linewidth]{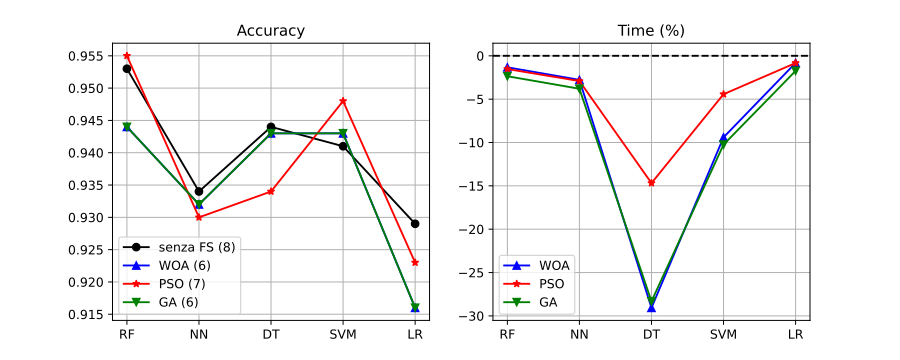}
  \caption{Accuracy and Time on dataset Diabete.}
  \label{fig:risultati_fs_d}
\end{figure}

Table \ref{tab:riassunto_prestazioni_d} summarizes the obtained accuracy.

\begin{table}[H]
  \centering
  \begin{tabular}{llllll}
    \toprule
      &\textbf{RF} & \textbf{NN}  &\textbf{DT} & \textbf{SVM} &\textbf{LR}\\
    \midrule
    No FS  &  95.4\%        &93.5\%           &94.4\%           & 94.1\% & 92.9\% \\
    WOA/GA &   94.4\% & 93.4\% & 94.3\% & \textbf{94.3\%} & 91.6\%\\
    PSO & \textbf{95.5\%} & 93\% & 93.4\% & \textbf{94.8\%} & 92.4\%\\
   
    \bottomrule
  \end{tabular}
  \caption{Accuracy}
  \label{tab:riassunto_prestazioni_d}
\end{table}

\begin{figure}[h]
  \centering
  \includegraphics[width=1\linewidth]{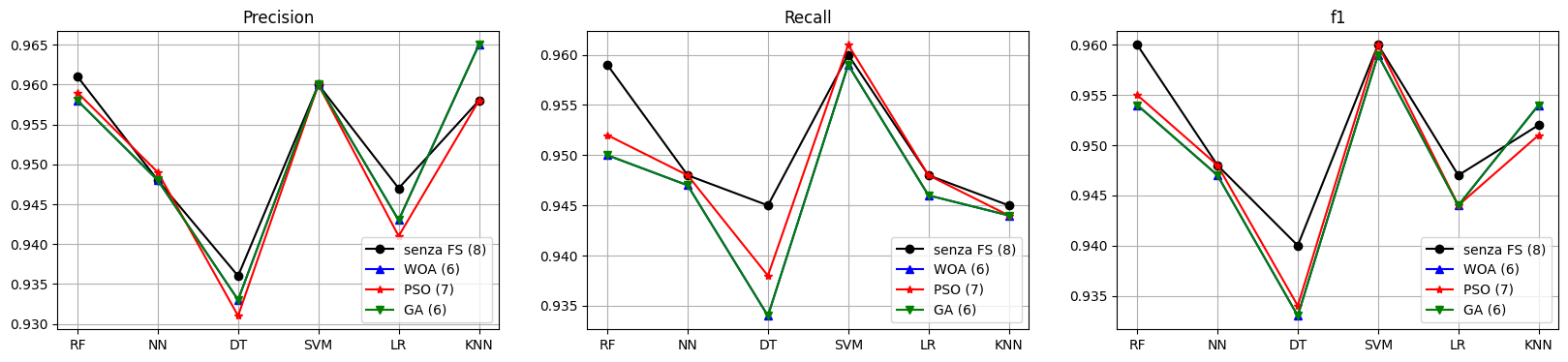}
  \caption{Precision, Recall and F1-score on dataset Diabete.}
  \label{fig:acc_recall_f1_diabete}
\end{figure}

In this dataset, the performance of WOA and GA with 2 fewer features than the full set decreased by a maximum of 1.3\%. The PSO although obtained the lowest fitness obtained slightly higher accuracies up to 0.5\%. Such a low increase in accuracy is not considered valid. Training times are all decreased by up to 30\%. Further metrics are shown in figure \ref{fig:acc_recall_f1_diabete}. 

\section{Discussions}
\label{sect:disc}
Summing up the results on the accuracy obtained we can see from the table \ref{tab:riassunto_acc} that:
\begin{itemize}
    \item The models most influenced by FS are RF, DT and LR.
    \item In the Heart Failure dataset, in which there were only 103 data points, the performance of the FS was disastrous. In this dataset, only the GA managed to increase the accuracy.
    \item In all datasets except Heart Failure, the FS decreased the training times of all models without significantly affecting performance. The highest accuracy increase value is +2.8\% for the decision tree with WOA in the breast cancer dataset, in which the feature reduction was 60\% and the training time decreased by 60\%.
     The lowest one is -4\% for linear regression with WOA in kidney disease, where the decrease in features was 83.3\% and the training time decreased by about 5\%. However, a small reduction in accuracy also occurs in the other FS algorithms.
    \item The genetic algorithm obtained accuracy $\geq$ of the complete set of features in 60\% of cases.
    \item The PSO obtained accuracy $\geq$ of the complete set of features in 50\% of cases.
    \item The WOA achieved accuracy $\geq$ of the complete set of features in 40\% of cases.
    \item In general, except for the Heart Failure dataset, the accuracies were always above 91\%.
    
\end{itemize}

Regarding the training times we note from the table \ref{tab:riassunto_tempi} that:
\begin{itemize}
    \item The most influenced models were DT and NN. Of these, the neural network performed worse in training time in 40\% of cases. This happens because the reduction of features makes it more complex for the neural network to optimize the generalization function.
    \item Overall, training times decreased for all models except the neural network, where the most significant increase was 100\%.
     \item the variations in the times of the $\pm 2 \%$ cannot be considered significant, as they may fall within the fluctuation range caused by variable hardware conditions and other software factors.
    \item Times remained unchanged or increased in 13\% of cases. 
\end{itemize}

\begin{table}[h]
\centering
\caption{Table of Models Accuracy.}
\label{tab:riassunto_acc}
\begin{tabular}{|c|c|c|c|c|c|c|}
\hline
\cline{2-7}
& & \textbf{Breast Cancer} & \textbf{Pime Indian} & \textbf{Heart Failure} & \textbf{Kidney Disease} & \textbf{Diabetes}\\
\hline
\multirow{4}{*}{RF} 
& No FS & 99.1\%                & 93.9\%  & 86\%         & 98\% & 95.4\%  \\
\cline{2-7}
& WOA & \textbf{99.3\%} (+0.2) & 93.9\%   & 71\%  (-15.0)& 98\% & 94.4\% (-1.0)   \\
\cline{2-7}
& PSO & \textbf{99.6\%} (+0.5)  & 93.9\%   & 69\% (-17.0)& \textbf{99.8\%} (+0.8) & \textbf{95.5\%} (+0.1) \\
\cline{2-7}
& GA & \textbf{99.4\%} (+0.3) & 93.9\% & \textbf{87\%} (+1) & \textbf{99.8\%}(+0.8) & 94.4\% (-1.0)\\
\hline

\multirow{4}{*}{NN} 
& No FS &  98.6\%               & 91\% & 76\%                 & 98.8\%              & 93.5\%  \\
\cline{2-7}
& WOA    & 97.6\% (-1.0)        & 91\% & 75\%(-1.0)           & 97.1\% (-1.7)       & 93.4\% (-0.1) \\
\cline{2-7}
& PSO &  \textbf{99.7\%} (+1.1) & 91\% & 46\% (-30.0)         & 97.1\% (-1.7)       & 93.0\% (-0.5)  \\
\cline{2-7}
& GA & \textbf{99.4\%} (+0.8)   & 91\% & \textbf{81\%} (+5.0) & \textbf{99.9\%} (+1.1) & 93.4\% (-0.1) \\
\hline

\multirow{4}{*}{DT} 
& No FS & 96\%                  & 89.9\%                  & 81\%        & 97.9\% & 94.4\%   \\
\cline{2-7}
& WOA & \textbf{98.8\%} (+2.8) & \textbf{90.1\%} (+0.2)  & 64\% (-17.0) & \textbf{99.1\%} (+1.2) & 94.3\% (-0.1) \\
\cline{2-7}
& PSO & \textbf{97.4\%} (+1.4) & \textbf{90.1\%} (+0.2) & 75\% (-6.0)& \textbf{98.1\%} (+0.2) & 93.4\% (-1.0) \\
\cline{2-7}
& GA & \textbf{97.4\%} (+1.4) & \textbf{90.1\%} (+0.2)  & 80\% (-1.0) & 97.1\% (-0.8) & 94.3\%  (-0.1)\\
\hline

\multirow{4}{*}{SVM} 
& No FS & 99.5\%        & 95\%              & 78\%        & 98.6\% & 94.1\%   \\
\cline{2-7}
& WOA &  98.4\% (-1.1) & 94\% (-1.0)         & 54\% (-24)& 98.6\% & \textbf{94.3\% } (+0.2) \\
\cline{2-7}
& PSO & \textbf{100\%} (+0.5) & 94\% (-1.0)  & 68\% (-10) & 98.6\% & 92.9\% (-1.2)\\
\cline{2-7}
& GA & \textbf{100\%} (+0.5) & 94\% (-1.0) & 75\% (-3) & 98.6\% & \textbf{94.3\%} (+0.2)  \\
\hline

\multirow{4}{*}{LR} 
& No FS & 100\%         & 88.1\%                & 68\%      & 98.6\%            & 92.9\%  \\
\cline{2-7}
& WOA &  97.9\% (-2.1) & \textbf{88.8\%} (+0.7) & 46\% (-22)& 94.6\% (-4.0)     & 91.6\% (-1.3)\\
\cline{2-7}
& PSO & 99.5\% (-0.5) & \textbf{88.8\%} (+0.7)  & 46\% (-22)& \textbf{100\%} (+1.4) & 92.4\% (-0.5) \\
\cline{2-7}
& GA & 98.9\%  (-1.1)& \textbf{88.8\%} (+0.7) & \textbf{71\%} (+3.0)& 94.6\% (-4.0) & 91.6\% (-1.3)\\
\hline

\multirow{4}{*}{KNN} 
& No FS & 98.7\%         & 90.7\%                & 70\%      & 96.1\%            & 95.1\%  \\
\cline{2-7}
& WOA &  98.1\% (-0.6) & \textbf{91.1\%} (+0.4) & 57.5\% (-12.5)& 95.4\% (-0.7)     & \textbf{95.4\%} (+0.3)\\
\cline{2-7}
& PSO & 98.5\% (-0.2) & \textbf{91.1\%} (+0.4)  & 65\% (-15)& \textbf{98.5\%} (+2.4) & 95\% (-0.1) \\
\cline{2-7}
& GA & 98.4\%  (-0.3)& \textbf{91.1\%} (+0.4) & \textbf{77.5\%} (+7.5)& 97.1\% (+1.0) & \textbf{95.4\%} (+0.3)\\
\hline

\end{tabular}
\end{table}

\begin{table}[h]
\centering
\caption{Table of Models Processing Time.}
\label{tab:riassunto_tempi}
\begin{tabular}{|c|c|c|c|c|c|c|}
\hline
\cline{2-7}
& & \textbf{Breast Cancer} & \textbf{Pime Indian} & \textbf{Heart Failure} & \textbf{Kidney Disease} & \textbf{Diabete}\\
\hline
\multirow{4}{*}{RF} 
& WOA & \textbf{-5\%} & \textbf{-2\%}   & \textbf{-1\%} & \textbf{-2\%} & \textbf{-2\%}   \\
\cline{2-7}
& PSO & \textbf{-3\%}  & \textbf{-6\%}   & \textbf{-1\%} & 0 & \textbf{-2\%} \\
\cline{2-7}
& GA & \textbf{-5\%} & \textbf{-3\%} & \textbf{-1\%} & 0 & \textbf{-3\%}\\
\hline

\multirow{4}{*}{NN} 
& WOA    &  +3\%       & \textbf{-6\%} & +100\% & +38\% & \textbf{-3\%} \\
\cline{2-7}
& PSO &  +3\% & \textbf{-4\%} & \textbf{-60\%} & +28\% & \textbf{-3\%}  \\
\cline{2-7}
& GA & \textbf{-12\%}  & \textbf{-3.9\%} & \textbf{-20\%} & +13\% & \textbf{-4\%} \\
\hline

\multirow{4}{*}{DT} 
& WOA & \textbf{-55\%} & \textbf{-9\%} & \textbf{-12\%} & \textbf{-15\%} &\textbf{-28\%} \\
\cline{2-7}
& PSO & \textbf{-45\%} & \textbf{-8\%} & \textbf{-20\%}& \textbf{-9\%} & \textbf{-14\%} \\
\cline{2-7}
& GA & \textbf{-55\%} & \textbf{-10\%} & \textbf{-20\%} & \textbf{-15\%} &\textbf{-28\%}\\
\hline

\multirow{4}{*}{SVM} 
& WOA &  \textbf{-14\%} & \textbf{-2\%}  & \textbf{-5\%} & \textbf{-8\%} & \textbf{-9\% } \\
\cline{2-7}
& PSO & \textbf{-13\%}& \textbf{-2\%} & \textbf{-5\%} & \textbf{-8\%} & \textbf{-9\%}\\
\cline{2-7}
& GA & \textbf{-14\%}& \textbf{-1\%} & \textbf{-5\%} & \textbf{-8\%} & \textbf{-9\%}  \\
\hline

\multirow{4}{*}{LR} 
& WOA &  \textbf{-25\%} & 0 & \textbf{-4\%} & \textbf{-5\%} & \textbf{-3\%}\\
\cline{2-7}
& PSO & \textbf{-43\%} & +1  & \textbf{-5\%} & \textbf{-5\%} & \textbf{-3\%} \\
\cline{2-7}
& GA & \textbf{-43\%} & \textbf{-1\%} & \textbf{-5\%} & \textbf{-5\%} & \textbf{-4\%}\\
\hline

\end{tabular}
\end{table}

The overall results for precision show that:
\begin{itemize}
    \item WOA had accuracy greater than or equal to that without FS in 36.67\% of cases.
    \item PSO had accuracy greater than or equal to that without FS in 50\% of cases.
    \item GA had accuracy greater than or equal to that without FS in 53.3\% of cases.
\end{itemize}

The overall results for recall show that:
\begin{itemize}
    \item WOA had recall greater than or equal to that without FS in 20\% of cases.
    \item PSO had recall greater than or equal to that without FS in 43\% of cases.
    \item GA had the recall greater than or equal to that without FS in 40\% of cases.
\end{itemize}

The overall results for the f1 score show that:
\begin{itemize}
    \item WOA had f1 greater than or equal to that without FS in 23\% of cases.
    \item PSO had f1 greater than or equal to that without FS in 33\% of cases.
    \item GA had f1 greater than or equal to that without FS in 50\% of cases.
\end{itemize}

The overall best-performing algorithm was the genetic algorithm. Overall performance, except for the heart failure dataset, was all above 87\%.

\section{Conclusion}
\label{sect:conclusion}

Our experiments highlight the significance of Feature Selection in enhancing machine learning model performance \cite{IntellisysLetteriCDP21} \cite{abs-2012-15231}. The impact varies based on factors such as the chosen FS algorithm, dataset characteristics, and model features \cite{eurospLetteriPC19}. 

It is worth noting that, employing FS yields superior results when applied to datasets with numerous features ($\geq$ 10).

The results show a prominent finding was a 2.8\% accuracy boost in the breast cancer dataset by reducing features from 30 to 12 the number of features, concurrently cutting training time by 55\%.

Reducing the number of features does not consistently reduce neural network training time. Fewer features often require more cycles for convergence demonstrating cycles and training times in the heart failure dataset, where performance was poorer. The experiments employed a neural network with 2 hidden layers of 10 neurons each, but feature reduction might necessitate a different network configuration.

Another critical observation involves the Particle Swarm Optimization (PSO) algorithm's challenge in improving fitness across generations. Graphs 4.1, 4.3, 4.5, 4.7, and 4.9 indicate PSO's struggle in achieving fitness improvements, possibly due to suboptimal initialization of algorithm parameters (w, c1, c2).

Furthermore, overall performance was consistently lower with metrics other than accuracy, due to the fitness function solely considering accuracy.

Furthermore, performance was generally lower with metrics other than accuracy, this is because the fitness function considered only accuracy as a metric.

\begin{table}[H]
  \centering
  \begin{tabular}{lllc}
    \toprule
    \textbf{Algorithm} & \textbf{Cycles} & \textbf{Time} & \textbf{\#Features}\\
    \midrule
no FS &  939 &  502.96 & 12 \\
WOA  & 1838 &  961.93 &  5\\
PSO &  280 &  145.29 &  2\\
GA & 967 &  503.44 &  3\\
    \bottomrule
  \end{tabular}
  \caption{The average number of cycles and time of training for a neural network on the heart failure dataset.}
  \label{tab:nn_time_epoch}
\end{table}
In general, bio-inspired algorithms are flexible and can be applied to a wide range of optimization problems, including problems that are highly nonlinear, non-convex, or have multiple objectives. Since they are gradient-free, they can find better solutions faster and are computationally cheaper than gradient-based methods. However, bio-inspired methods do not guarantee convergence to a global optimum. The key aspect for bio-inspired methods to perform well is to balance the exploration and exploitation phases, which typically require parameter tuning to achieve optimal performance. This can be time-consuming and requires expertise in the algorithms being used. Further tuning of the parameters of the bio-inspired algorithms used for FS is subject to our future work. In addition, we would like to experiment with additional new fitness criteria, for example, changing the alpha parameter to evaluate performances when more importance is given to reducing features rather than accuracy. Another future work is to experiment with different types of structured and unstructured data. Finally, we would like to mention that this work is part of our research project on healthcare assistant agents, in which we are working on different aspects of these agents, including but not limited to, ethical aspects (\cite{digitalsociety/DyoubCL22}, \cite{DyoubCL22}, \cite{DyoubCLL20}), and considering data storing and management using Multi-agent systems combined with blockchains \cite{abs-2311-01584} and a Reinforcement Learning approach logic-based driven \cite{de2023extension}.


%
%
%
\bibliographystyle{splncs04}
\bibliography{bio}

\end{document}